\documentclass[review]{elsarticle}

\usepackage{lineno,hyperref,booktabs,multirow,xcolor}
\usepackage{dingbat}
\usepackage{lscape}
\usepackage{geometry}
\usepackage{rotating}
\usepackage{graphicx}
\usepackage{makecell}
\usepackage{caption}
\usepackage{tablefootnote}
\biboptions{authoryear}
\usepackage{threeparttable}
\usepackage{tabularx, makecell}
\begin{document}
\begin{frontmatter}
\title{\textit{Where is VALDO?} VAscular Lesions Detection and segmentatiOn challenge at MICCAI 2021}


\author[1,2,3]{Carole H. Sudre\fnref{mycorrespondingauthor}\corref{contrib}}

\fntext[mycorrespondingauthor]{Corresponding author:c.sudre@ucl.ac.uk}
\cortext[contrib]{Authors contributed equally to this work}
\author[4]{Kimberlin Van Wijnen\corref{contrib}}

\author[4]{Florian Dubost} 
\author[5]{Hieab Adams} 
\author[6]{David Atkinson} 
\author[2,7]{Frederik Barkhof} 
\author[4]{Mahlet A. Birhanu} 
\author[4]{Esther E. Bron} 
\author[4]{Robin Camarasa}
\author[1]{Nish Chaturvedi }
\author[9]{Yuan Chen} 
\author[10]{Zihao Chen }
\author[4]{Shuai Chen }
\author[11]{Qi Dou} 
\author[5]{Tavia Evans} 
\author[12,24]{Ivan Ezhov} 
\author[13]{Haojun Gao} 
\author[14]{Marta Girones Sanguesa} 
\author[15,28,29]{Juan Domingo Gispert} 
\author[16]{Beatriz Gomez Anson} 
\author[1]{Alun D. Hughes} 
\author[17]{M. Arfan Ikram} 
\author[7]{Silvia Ingala} 
\author[18]{H. Rolf Jaeger} 
\author[12,23,24]{Florian Kofler} 
\author[14]{Hugo J. Kuijf} 
\author[14]{Denis Kutnar} 
\author[]{Minho Lee} 
\author[4]{Bo Li} 
\author[7]{Luigi Lorenzini} 
\author[12,25]{Bjoern Menze} 
\author[15,27]{Jose Luis Molinuevo} 
\author[19]{Yiwei Pan} 
\author[20]{Elodie Puybareau} 
\author[18]{Rafael Rehwald}
\author[4]{Ruisheng Su}
\author[19]{Pengcheng Shi} 
\author[]{Lorna Smith} 
\author[1]{Therese Tillin} 
\author[20]{Guillaume Tochon} 
\author[21]{H\'el\`ene Urien} 
\author[14]{Bas H.M. van der Velden} 
\author[8,17]{Isabelle F. van der Velpen} 
\author[23]{Benedikt Wiestler} 
\author[8,17]{Frank J. Wolters} 
\author[17]{Pinar Yilmaz} 
\author[4,26]{Marius de Groot} 
\author[8,17]{Meike W. Vernooij} 
\author[4,22]{Marleen de Bruijne} 
\author{for the ALFA study}\fnref{groupstudy}
\fntext[groupstudy]{The complete list of collaborators for the ALFA study can be found in acknowledgments}

\address[1]{MRC Unit for Lifelong Health and Ageing at UCL, University College London, London,  United Kingdom}

\address[2]{Centre for Medical Image Computing, University College London,  London, United Kingdom}

\address[3]{School of Biomedical Engineering and Imaging Sciences, King's College London,  London,  United Kingdom}

\address[4]{Biomedical Imaging Group Rotterdam, Department of Radiology and Nuclear Medicine, Erasmus MC, Rotterdam, The Netherlands}

\address[5]{Department of Clinical Genetics and Radiology, Erasmus MC, Rotterdam, The Netherlands}

\address[6]{Centre for Medical Imaging, University College London, London United Kingdom}

\address[7]{Department of Radiology and Nuclear Medicine, Amsterdam University Medical Centre, Amsterdam, The Netherlands}

\address[8]{Department of Radiology and Nuclear Medicine, Erasmus MC, Rotterdam, The Netherlands}

\address[9]{Department of Radiology, University of Massachusetts Medical School, Worcester, The USA}

\address[10]{School of Biomedical Engineering, Shanghai Jiao Tong University, Shanghai, China}

\address[11]{Department of Computer Science and Engineering, The Chinese University of Hong Kong, China}

\address[12]{Department of Informatics, Technische Universitat Munchen, Munich, Germany}

\address[13]{Department of Radiology, Zhejiang University, Hangzhou, China}

\address[14]{Image Sciences Institute, University Medical Center Utrecht, Utrecht, the Netherlands}
\address[15]{Barcelona$\beta$ Brain Research Center (BBRC), Pasqual Maragall Foundation, Barcelona, Spain}
\address[16]{Department of Radiology, Hospital San Pau i santa Creu, Barcelona, Spain}

\address[17]{Department of Epidemiology, Erasmus MC, Rotterdam, The Netherlands}

\address[18]{Institute of Neurology, University College London, London, United Kingdom}

\address[19]{Department of Electronic and Information Engineering, Harbin Institute of Technology at Shenzhen, Shenzhen, China}

\address[20]{LRDE, EPITA, Paris, France}

\address[21]{ISEP-Institut Supérieur d’Électronique de Paris, Issy-les-Moulineaux, France}

\address[22]{Department of Computer Science, University of Copenhagen, Copenhagen, Denmark}

\address[23]{Department of Diagnostic and Interventional Neuroradiology, School of Medicine, Klinikum rechts der Isar, Technical University of Munich, Germany}

\address[24]{TranslaTUM - Central Institute for Translational Cancer Research, Technical University of Munich, Germany}

\address[25]{Department of Quantitative Biomedicine, University of Zurich, Switzerland}

\address[26]{GlaxoSmithKline Research, Stevenage,UK}

\address[27]{H. Lundbeck A/S, Copenhagen, Denmark}
\address[28]{Hospital del Mar Medical Research Institute (IMIM), Barcelona, Spain}
\address[29]{Centro de Investigación Biomédica en Red Bioingeniería, Biomateriales y Nanomedicina, (CIBER-BBN), Barcelona, Spain}


\begin{abstract}Imaging markers of cerebral small vessel disease provide valuable information on brain health, but their manual assessment is time-consuming and hampered by substantial intra- and interrater variability. Automated rating may benefit biomedical research, as well as clinical assessment, but diagnostic reliability of existing algorithms is unknown. Here, we present the results of the \textit{VAscular Lesions DetectiOn and Segmentation} (\textit{Where is VALDO?}) challenge that was run as a satellite event at the international conference on Medical Image Computing and Computer Aided Intervention (MICCAI) 2021.
This challenge aimed to promote the development of methods for automated detection and segmentation of small and sparse imaging markers of cerebral small vessel disease, namely enlarged perivascular spaces (EPVS) (Task 1), cerebral microbleeds (Task 2) and lacunes of presumed vascular origin (Task 3) while leveraging weak and noisy labels. Overall, 12 teams participated in the challenge proposing solutions for one or more tasks (4 for Task 1 - EPVS, 9 for Task 2 - Microbleeds and 6 for Task 3 - Lacunes). Multi-cohort data was used in both training and evaluation. Results showed a large variability in performance both across teams and across tasks, with promising results notably for Task 1 - EPVS and Task 2 - Microbleeds and not practically useful results yet for Task 3 - Lacunes. It also highlighted the performance inconsistency across cases that may deter use at an individual level, while still proving useful at a population level. 

\end{abstract}






\end{frontmatter}


\section{Introduction}\label{sec1}

Cerebral small vessel disease (CSVD), the deterioration of the smallest brain vessels, encompasses a large variety of etiologies including arteriolosclerosis \citep{alistair2002hypertensive} and amyloid pathology \citep{kester2014associations} and may be further driven by genetic predisposition \citep{haffner2016genetic,giau2019genetic}. It results in observable damage or changes to the brain. Most commonly observed MRI markers of CSVD include white matter hyperintensities (WMH), cerebral microbleeds, lacunes of presumed vascular origin, and enlarged perivascular spaces  \citep{wardlaw2013neuroimaging}.
CSVD related damage has been associated with an increased risk of stroke and dementia, and with the acceleration of cognitive decline \citep{ostergaard2016cerebral,rensma2018cerebral}. The presence of these markers are also associated to one another \citep{zhang2014risk,zhu2010severity,yates2014cerebral}.

 WMH are the most visible marker of CSVD and have naturally taken the centre stage of clinical research in CSVD. In addition, research on development of WMH segmentation solutions has been particularly popularized thanks to impactful research showing the clinical importance of lesion volumetry \citep{van2006impact}. While the automated quantification of white matter hyperintensities has been heavily studied for the last decade with very successful solutions \citep{sudre2015bayesian,guerrero2018white,atlason2019segae,de2009white}, automated detection and segmentation of the small, focal markers of CSVD has been investigated less frequently. However, as the interest of the clinical community in these markers starts to grow, getting to understand their relevance in clinical research requires them to be adequately detected and quantified. While these markers are currently typically assessed visually through binary dichotomization (presence vs absence) \citep{yates2014cerebral}, counts \citep{adams2015priori}, or visual scales \citep{potter2011neuroimaging}, such visual assessment is time consuming and subject to large inter- and intra-rater variability \citep{sudre2019let}. Automated methods are therefore required to make quantification robust and reliable as well as feasible in the context of large data sets. So far, development of automated methods has been impeded by the methodological issues related to the very small size of these markers and the resulting extreme imbalance in the data, as well as the absence of a gold standard for annotation.

Methodological developments towards automated solutions for the quantification of biomarkers have found a new dynamic thanks to the annotated datasets made available through technical challenges on segmentation and detection in brain MRI with notably the popular BRATS challenge \citep{menze2014multimodal}, ISLes \citep{maier2017isles}, MRBrainS \citep{mendrik2015mrbrains}, the 2017 MICCAI WMH challenge \citep{kuijf2019standardized} or the more recent ADAM challenge \citep{TIMMINS2021118216} on intracranial aneurysms. Such challenges give insight into state-of-the-art methodology and remaining technical problems for a specific question.

The \textit{VAscular Lesions DetectiOn and Segmentation} (\textit{Where is VALDO?}) challenge was organized with the aim of promoting the development of new solutions for the automated detection and segmentation of these sparse and small structural brain changes (enlarged perivascular spaces (Task 1), cerebral microbleeds (Task 2) and lacunes (Task 3) ) while leveraging weak and noisy labels from manual annotation or visual assessment. Beyond a simple benchmarking exercise assessing the state of the solution space, this challenge was further intended to gain insight on the current pitfalls and challenges, raise awareness and contribute to the building of a community dedicated to developing solutions to facilitate quantification of CSVD markers in brain MRI scans. This paper describes the design, results, and lessons learnt through the challenge according to the reporting guidelines detailed in \citep{maier2020bias}.    
\section{Methods}
\subsection{Mission of the challenge}

\begin{figure}
    \includegraphics[angle=90,width=0.95\textwidth]{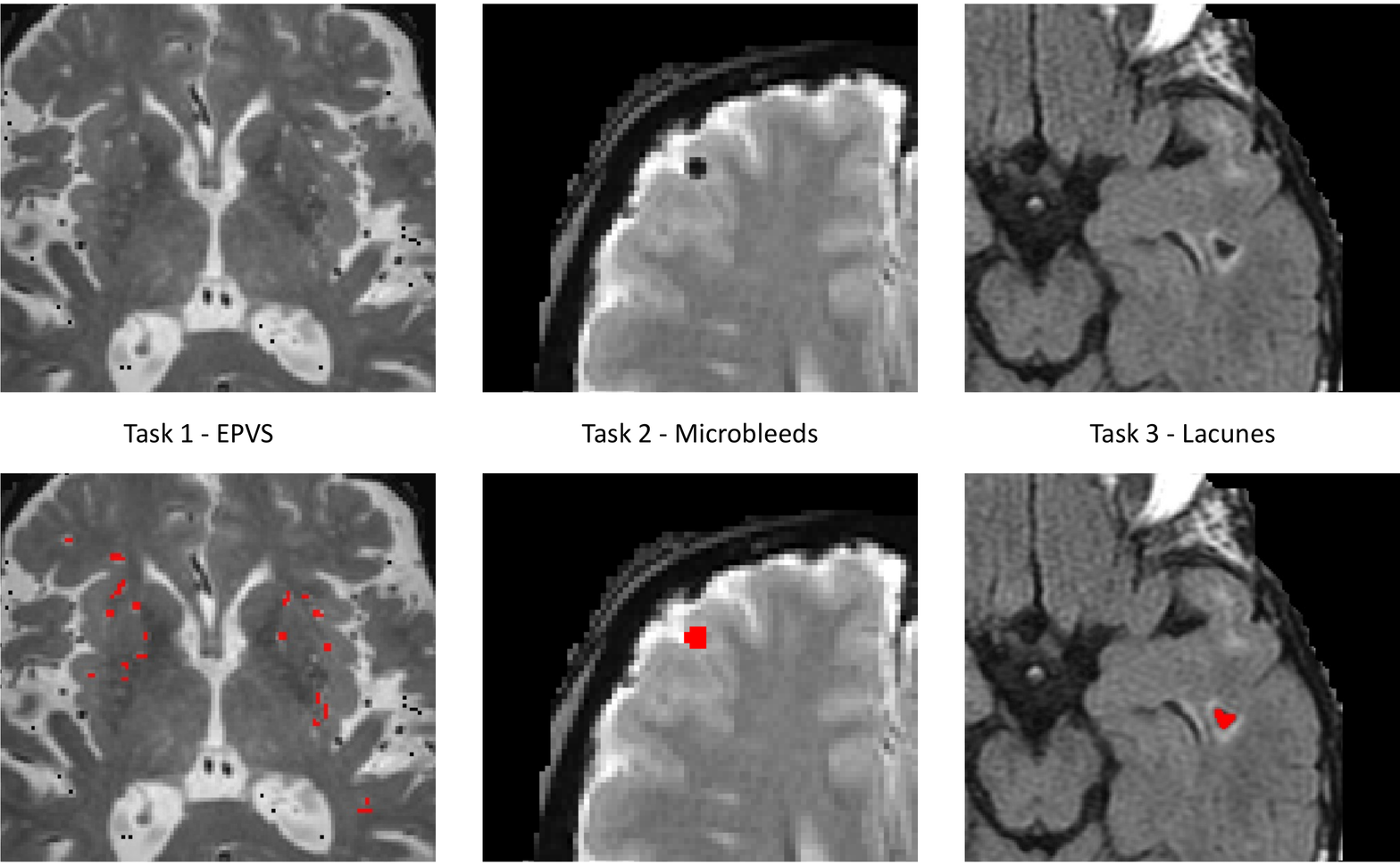}
    \caption{Annotated example of the three type of markers targeted in the challenge}
    \label{fig:example_annot}
\end{figure}

The \textit{Where is VALDO?} challenge was organized to assess three tasks, each of them focusing on one focal marker of CSVD - Task 1 on enlarged perivascular spaces (EPVS), Task 2 on cerebral microbleeds and Task 3 on Lacunes. Figure \ref{fig:example_annot} illustrates each of these markers as annotated in the challenge training set. 

Currently, the lack of accurate and reproducible automated methods for all three markers prohibits the identification of clinically relevant characteristics at both individual and population levels. Therefore, for each of the stated markers both detection and segmentation performance need to be assessed. Ultimately, the improved quantification of these small focal markers of CSVD may be used to better understand their relevance and derive biomarkers for diagnosis or prognosis in the context of healthy ageing and dementia, and as surrogate end points in clinical trials. 

In proposing tasks particularly subject to high data imbalance and limited and/or noisy annotations, this challenge further aimed to catalyse methodological research to address these common issues in the medical image analysis community.


Ultimately, the proposed methods should be applicable to different settings involving ageing populations such as population cohorts, clinical trials or memory clinics. 

The challenge dataset however consisted exclusively of population-based cohorts - two to three according to the task, with differences in MRI acquisition protocol, image resolution and scanner characteristics across datasets. No additional information beyond the images was provided. Each of the datasets was enriched for lesion burden through stratified sampling of the skewed population distributions. 

For each task, a similar approach to assessment was adopted to ensure consistency across tasks and address both segmentation and detection aspects, although some may currently be considered more important in one task than another, with different paradigms used in clinical practice. For instance, the blooming effect observed in the presence of microbleeds is protocol dependent, making the detection more relevant than the segmentation in that task \citep{buch2017determination}.

\subsection{Challenge organization}
The \textit{Where is Valdo?} challenge was run as a satellite event at MICCAI 2021 as a collaboration of University College London and Erasmus MC University Medical Center Rotterdam. Its three-task design was peer-reviewed prior to acceptance and made public at \url{https://doi.org/10.5281/zenodo.4600654} 
 Regarding prize eligibility, it was decided that organizers would not participate and while members of the same institutions as the organizers were allowed to participate in the challenge, they would not be eligible for prizes. Prizes were given to each winner of individual tasks and the overall winner across all tasks. Results were publicly presented for all participating teams. All submitting teams were invited to propose two team members (per task) to participate as co-authors in the challenge overview paper. After publication of this overview paper of the challenge, the submission will reopen to the community for anyone wanting to benchmark their methods against those previously submitted. Further information is available on the challenge website \url{https://valdo.grand-challenge.org}.

\begin{figure}
    \includegraphics[width=0.95\textwidth]{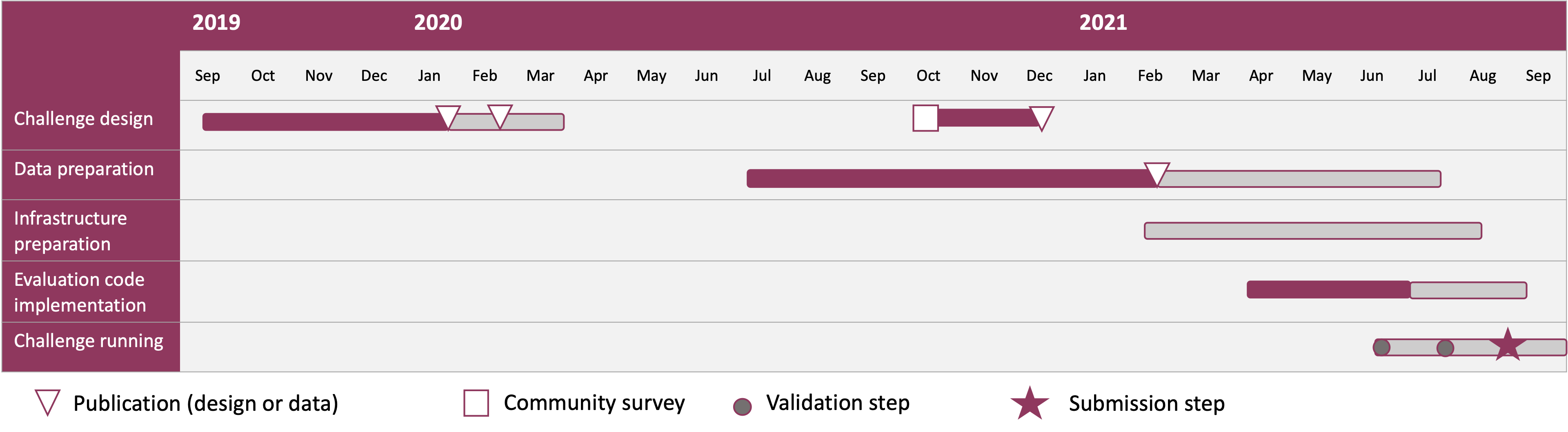}
    \caption{Timeline of the challenge from inception in September 2019}
    \label{fig:organization}
\end{figure}

The challenge was organized in 4 phases: 1) a training phase from the moment the annotated database was made downloadable (February 2021), 2\&3) two optional validation steps on 5 new cases to provide individual (no public leaderboard) feedback on the performance (14\textsuperscript{th} to 21\textsuperscript{st} of June and 12\textsuperscript{th} to 19\textsuperscript{th} of July) and 4) the final evaluation stage on withheld cases (submission from 26\textsuperscript{th} of July to 5\textsuperscript{th} of August 2021). A grace period extending until the 10\textsuperscript{th} August in case of technical difficulties was granted to all participants. Participants had to provide a docker container for their fully automated method (1 for each task) and were allowed to participate in any or all the tasks. Use of additional training data was allowed under the condition it would be made available at submission time. The methods did not have to be similar across all tasks. Details of the submission procedure are listed at \url{https://valdo.grand-challenge.org/Submission/}. Participating teams were also requested to provide a short technical note describing their solutions that have been made available at \url{https://openreview.net/group?id=MICCAI.org/2021/Challenge/VALDO}. Figure \ref{fig:organization} presents the timeline of the challenge.

Submitted data were evaluated on the test set at a GPU facility at Erasmus MC. In order to ensure that the proposed methods were running as expected, each docker was run on one example of the training set and the result sent back to the participants for checking, allowing for submission of a new docker if the output was not as expected. 

The evaluation code was made available prior to submission at \url{https://github.com/WhereIsValdo/valdo-eval-2021}. The participating teams were encouraged to make their source code publicly available and all participants except one team agreed for their docker containers to be made public. They have been placed on \url{https://hub.docker.com/r/whereisvaldo/challenge2021/tags}

The challenge was sponsored by NVIDIA and Icometrix. Test data was available to CHS and KVW. The contribution of the authors listed in this manuscript can be found in supplementary material.

\subsection{Community survey}
To better understand the interest within the community for such initiative, we launched in January 2021 a survey targeting the community working in the field of automated detection of CSVD lesions. This survey was sent to a list of researchers having recently published automated methods for detection or segmentation of one of the three lesion types considered in the \textit{Where is VALDO?} challenge, the International Society of Vascular Behavioural and Cognitive Disorders (VasCog \url{https://www.vas-cog.com}), and the Medical Image Understanding and Analysis (MIUA \url{miua@jiscmail.ac.uk}) mailing list, and the survey was shared on social media by the challenge organizing team. Overall, 36 answers were recorded with 25 individuals indicating to be very likely or likely to participate. Among the respondents, 39\% indicated being already actively working in the field of CSVD and 30\% more general in the neuroimaging field. Microbleed segmentation appeared as the most popular task in the survey with 15 respondents indicating they were highly likely to participate in this task against 10 for EPVS and 10 for lacunes. These answers helped shape the final challenge design, notably standardizing the evaluation of the different tasks and making the challenge overall more concise.

\subsection{Challenge data sets}
The challenge data sets (training, validation, and test sets) came from the same cohorts with a similar ratio between them across tasks. This ratio was also kept in the testing set. 

\subsubsection{Datasets and image acquisition}
Two subsets of population cohorts were used for all three tasks and an additional one was further available for the microbleed detection/segmentation task, namely the SABRE and Rotterdam Scan Study (RSS) cohorts and the ALFA study respectively. All cohorts were retrospective studies for which local ethical approval had already been obtained from the National Research Ethics Service Committee, London-Fulham (14/LO/0108) for SABRE, the Population Research Act from the Ministry of Health for RSS and the Independent Ethics Committee Parc de Salut Mar Barcelona and registered at Clinicaltrials.gov (NCT01835717) for ALFA. For all datasets, acquisition of the data was performed by a trained radiographer according to a predefined research protocol. The training data for the \textit{Where is VALDO?} challenge was made available under a CC BY NC-SA license. 

\paragraph{SABRE}
The Southall and Brent Revisited (SABRE) cohort is a population cohort of individuals residing in the two named boroughs of west London (UK)\citep{tillin2013relationship}. This tri-ethnic cohort was initially recruited in 1988 with the purpose of investigating metabolic and cardiovascular diseases across ethnicities. For their third clinical visit (2014-2018), life partners were also invited to take part and study participants underwent a brain MRI session on a Philips 3T scanner. Mean age in this cohort at time of acquisition was 72 years old ranging from 36 to 92.

\paragraph{RSS}
The Rotterdam Scan Study (RSS) \citep{ikram2015rss} is part of the larger Rotterdam Study (RS) \citep{ikram2020rotterdamstudy}, a population-based study that aims to investigate chronic illness in the elderly. Started in 1995, the Rotterdam Scan Study initially concerned a selection of the RS but since 2005 brain MRI is part of the core protocol of the study. Individuals aged 45 and over without dementia are eligible for MRI and are followed up every 3-4 years. Since 2005, scanning has been performed on a 1.5T GE MRI scanner dedicated to the study.  

\paragraph{ALFA}
The ALFA (Alzheimer's and Families) cohort is based on the ALFA registry that gathers details of relatives (generally offspring) of patients with Alzheimer's Disease making up for a cohort naturally enriched for genetic predisposition to AD. As described in the related protocol paper \citep{molinuevo2016alfa}, the ALFA cohort is composed of cognitively normal participants aged 45-74. Brain MRI sequences were acquired on a GE Discovery 3T scanner.

Table \ref{tab:acquisition} summarizes the acquisition parameters for the different sequences across the studied cohorts.

\begin{table}[hpb]
    \begin{tabular}{ccp{2cm}ccccc}
    \toprule
         Cohort&Sequence& Type & TR & TE & TI & FA & Resolution (mm) \\
         \midrule
         \multirow{4}{*}{SABRE}& T1w & Inversion prepared gradient echo & 6.9 & 3.1 & / & / & 1.09 x 1.09 x 1.0 \\
         & T2w & 3D sagittal turbo spin echo & 2500 & 222 & 836 & 8 & 1.09 x 1.09 x 1.0 \\
         & FLAIR & &4800 &125 &1650 & & 1.09 x 1.09 x 1.0   \\
         & T2* & Gradient echo & 1288 & 21 & / & 18 & 0.45 x 0.45 x 3.0 \\
         \midrule
         \multirow{4}{*}{RSS} & T1w & Gradient recalled echo & 13.8 & 2.8 & 400 & 20 & 0.49 x 0.49 x 0.8 \\
         & T2w & Fast spin echo & 12300 & 17.3 & / &/ & 0.49 x 0.49 x 0.8 \\
         & FLAIR & Fast spin echo & 8000 & 120 & 2000 & & 0.49 x 0.49 x 0.8\\
         & T2* & Gradient recalled echo & 45&31 &/ &13 & 0.49 x 0.49 x 0.8 \\
         \midrule
         \multirow{3}{*}{ALFA} & T1w & 3D & 8.0 &3.7 &450 &8 &1.0 x 1.0 x 1.0 \\
         & T2w & Fast spin echo & 5000 & 85 & / &110 & 1.0 x 1.0 x 3.0 \\
         & T2* & Gradient recalled echo & 1300 &23 & / & 15 & 1.0 x 1.0 x 3.0 \\
    \end{tabular}
    \caption{Acquisition details for the three cohorts. Acronyms FA - Flip angle; TE - echo time(ms); TI - inversion time(ms); TR - repetition time (ms)}
    \label{tab:acquisition}
\end{table}
\subsubsection{Training, validation and testing data}
For Task 1 - EPVS and Task 3 - Lacunes, imaging data consists of T1-weighted, T2-weighted and FLAIR images, with the latter two modalities rigidly registered to the T1 image using NiftyReg \citep{modat2014global}. For Task 2 - Microbleeds, imaging data is the combination of T2, T2* and T1-weighted images in T2* space.
Table \ref{tab:count_scans} presents the number of cases used for training and testing across the different tasks and the different cohorts. For each task, validation consisted of 5 cases from the RSS cohort. There was no overlap between training, test or validation datasets.
\begin{table}[]
    \centering
    \begin{tabular}{c|c|c|c|c|c|c|}
         & \multicolumn{2}{c|}{Task 1 - EPVS} & \multicolumn{2}{c|}{Task 2 - Microbleeds} & \multicolumn{2}{c|}{Task 3 - Lacunes}  \\
      Cohort   & Train & Test & Train & Test & Train &Test \\
      \hline
      SABRE & 6 & 10 & 11 & 20 & 6 & 10 \\
      RSS & 34 (6/28) & 56 & 34 & 68 & 34 & 56 \\
      ALFA & / & / & 27& 59 & / & / \\
      \hline
      Total  & 40 & 66 & 72 & 147 & 40 &66 \\
    \end{tabular}
    \caption{Number of cases in train and test set for each task and cohort origin. For RSS Task 1 of training separation between cases with full annotation and cases with only counts}
    \label{tab:count_scans}
\end{table}

The number of cases proposed for training was chosen based on annotation availability and data policy for making a certain number of cases publicly available. For Task 1 - EPVS and Task 3 - Lacunes, the SABRE segmentation data was already available for a set of 16 cases with high level of cerebrovascular damage. In comparison, for the RSS study, for which annotations were more widely available, data were selected to cover the variability in burden present in the study. They present close to a uniform distribution in burden thereby limiting data skewness towards cases without any lesion. In all tasks, annotated cases were distributed across training and testing set to follow approximately similar burden distribution. A ratio of 6:10 between training and testing data was chosen across all cohorts and tasks.

\subsubsection{Annotation}
Across the three cohorts, raters were all trained for their annotation task and had at least 3 years of professional experience in dealing with medical images. The segmentation was performed for all SABRE and ALFA cases using ITKSnap \citep{yushkevich2016itk}. For the RSS cases a custom MeVisLab \citep{ritter2011medical} application was used. In all cases were two annotations were available, the average of the two annotations was used as reference.
\paragraph{Task 1 - Enlarged Perivascular Spaces}
For Task 1, the annotation strategy differed between the SABRE and RSS cohort. For identifying EPVS, the STRIVE criteria \citep{wardlaw2013neuroimaging} for EPVS were used in the SABRE cohort, while in the RSS cohort, the UNIVRSE criteria \citep{adams2015priori} were used. These criteria are very similar, except for the fact that the UNIVRSE criteria only consider EPVS with a diameter between 1 and 3 mm, while the STRIVE criteria do not have a lower limit and consider any EPVS with a diameter up to 3 mm. 
In the SABRE cohort, EPVS over the whole brain image were annotated independently by two raters (CHS and LL) with a senior radiologist (BGA) confirming the segmentation of CHS. The three modalities were jointly used for the segmentation that was assessed across the three axes. 

For this dataset the annotation was provided in either of two forms: over the full brain or on only 5 randomly selected slabs of 5mm. A mask was provided per case indicating the slabs that were annotated.

In the RSS cohort, EPVS were annotated with segmentations in limited axial slices for 6 cases of the training set and the full test set, while the remaining 28 cases of the training set were annotated with dots only by a team of trained annotators supervized by KVW, FD and MWV. EPVS were annotated in four brain regions: the mesencephalon, hippocampus, basal ganglia, and the centrum semi-ovale. The first two smaller regions were annotated entirely. For the latter two regions, only one fixed slice was annotated. For the cases with EPVS segmentations, additional slices of the basal ganglia and of the white matter were annotated, the depth of these axial slices was randomly chosen per case. A mask indicating which parts of the brain had been annotated was computed using parcellation outputs for each case. 

\begin{figure}
    \centering
    \includegraphics[width=0.95\textwidth]{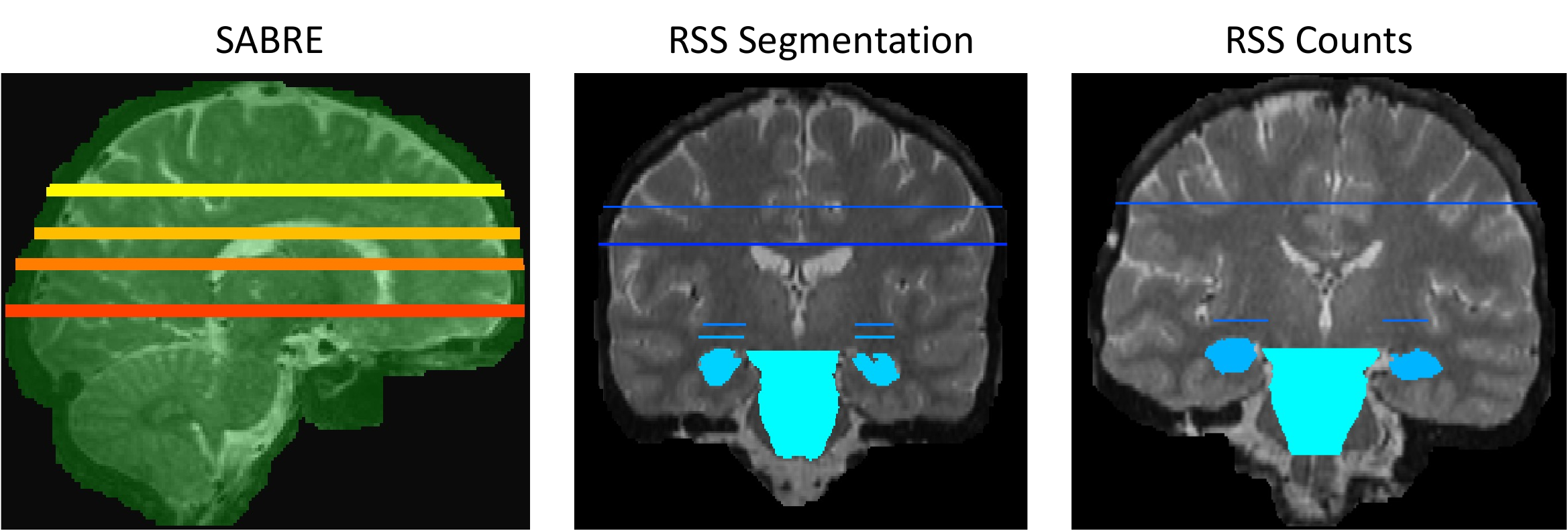}
    \caption{Example of annotation provided for Task 1 - EPVS with left) for SABRE slabs of 5 mm randomly selected or full segmentation over the image, middle) Segmentation on two slices of CSO, 2 slices of the basal ganglia, the hippocampi and mesencephalon for 6 RSS cases and right) count of EPVS on 1 slice of CSO, 1 slice of basal ganglia, hippocampi and mesencephalon for 28 cases of RSS. }
    \label{fig:Labelling}
    
\end{figure}

For the training data made available to participants, the EPVS annotations were either presented just as counts (computed from the dots), per slice and per region or as segmentations plus counts in the same areas. The masks  indicating the annotated regions and slices per case was also provided. Figure \ref{fig:Labelling} illustrates the type of annotation masks that were provided to the participants.


\paragraph{Task 2 - Microbleeds}
Different raters annotated each of the cohorts but followed very similar protocols. The BOMBS criteria \citep{cordonnier2009improving} was applied for the SABRE (RR under the supervision of HRJ) and ALFA cohort (consensus of SI and LL under the supervision of FB) as described in \citep{ingala2020relation}. A team of trained raters under the supervision of MWV applied the protocol described in \citep{vernooij2008prevalence} for RSS. Both identification protocols are in line with the STRIVE guidelines \citep{wardlaw2013neuroimaging} that indicate that microbleeds are areas of signal void of generally 2-5 mm in diameter but can be up to 10 mm. 

\paragraph{Task 3 - Lacunes}
Lacunes were identified using the STRIVE criteria \citep{wardlaw2013neuroimaging}. Cerebellar lacunes were excluded because of assumed differences in the underlying pathology in this brain region\citep{sigurdsson2022cerebrovascular}. Any surrounding gliosis (the hyperintense rim visible on FLAIR sequences) was not included in the segmentation of the lacune.
For the SABRE cohort, lacunes were identified at the same time as EPVS simply being assigned another label in the segmentation, with the two raters (CHS, LL) performing the identification and segmentation independently. For the RSS cohort, lacunes were independently segmented for all cases by two raters, the pair of raters varying across the cases. In RSS, all cases of training, validation and test set indicated by radiological reads as containing at least one lacune were consistently annotated by one rater (TE) on a custom MeVisLab\citep{ritter2011medical} application. The second set of annotations was performed using ITKSnap\citep{yushkevich2016itk}. PY annotated all cases of the training set. FW annotated the validation set as well as half of the test set. The remaining half of the test set was annotated by IFV. 

\subsubsection{Sources of annotation errors}
In all tasks, possible source of errors in the annotations pertain to multiple distinct sources: the appropriate identification of a target element either because these elements are very small and may be easy to miss or because it may be difficult to distinguish them from similarly appearing structures (mimics); the decision on the boundary of an object, probably notably more complex in a coarser resolution plane; the use of the segmentation software (too large brush, not considering all orientations for consistency or not adequately using the zoom). In the case of EPVS, identification of "large enough" marker was also a subjective consideration possibly leading to different detection levels.

\subsubsection{Preprocessing}
For all tasks, the preprocessing consisted of a rigid alignment of the images as indicated in section 2.4.2. A defacing mask derived from the T1-weighted image was applied to all registered modalities. While such a step would not be required in practice, this step was mandated by the data sharing policies around public release of the data. 
The defacing mask was obtained as the inverse of a dilated version of the brain mask as obtained from HD-BET \citep{isensee2019automated}. All RSS scans were corrected for intensity inhomogeneity with the default parameters of MINC N3 package \citep{sled1998nonparametric}.


\subsection{Assessment method}

All three tasks were evaluated using similar metrics in order to assess both detection and segmentation performance of the proposed solutions. A combination of relative error (F1 score and Mean Dice score) and absolute error (absolute element difference (AED) and absolute volume difference (AVD)) metrics was chosen, since they provide complementary information. The F1 score and the AED on the number of detected lesions were chosen as detection metrics while the Mean Dice score over the appropriately identified elements and the AVD were the metrics used for the evaluation of segmentation. Table \ref{tab:metrics} summarizes the purpose, formula and properties of the metrics used in the challenge across all tasks and calculated for each case, where $_c$ refers to 6-neighborhood connected components, TP to true positives, FP to False positive, FN to false negatives, Ref to the reference annotation and Seg to the predicted segmentation.
\begin{table}[]
    \centering
    \begin{tabular}{c|cccc}
    \toprule
      Metric   & Target & Formula& Range & Best\\
      \midrule
       F1 Score  & Detection & $\frac{2TP_{c}}{2TP_{c}+FP_{c}+FN_{c}}*100$&0 - 100&100\\
       AED & Detection & $\vert\#_{c}Ref-\#_{c}Seg\vert$& 0 - $\inf$& 0 \\
       Mean Dice & Segmentation & $\frac{100}{\#TP_{c}}\sum_{t\in TP_{c}} \frac{2 * \sum (Ref_{t} * Seg_{t}) }{\sum Ref_{t} + \sum Seg_{t}}$ & 0 - 100 & 100 \\
       AVD & Segmentation & $\vert Ref - Seg \vert$ & 0 - $\inf$ & 0\\
       \bottomrule
    \end{tabular}
    \caption{Description of detection and segmentation metrics used across all tasks for the evaluation.}
    \label{tab:metrics}
\end{table}

One essential aspect in the evaluation for the derivation of both F1 and Mean Dice score was the definition of true positive elements. To determine which of the elements were true positives, for all three tasks, connected components with a neighbourhood of 6 were established for both annotation and prediction using a threshold for the probability of 0.5 for the prediction map. Each annotation element was matched to at most one element from the prediction. For Task 1 - EPVS, a possible matchable element had to have an Intersection over Union (IoU) of more than 10\%. For Task 2 - Microbleeds and Task 3 - Lacunes, matching was possible when the centre of mass of the prediction element was less than 5 mm away from the center of mass of the ground truth segmentation element. When multiple elements were found to be matchable, the one with best association value (IoU or centre of mass distance) was attributed to the annotated label. For empty cases, the relative metrics were inapplicable, so only the absolute error metrics (number of elements and volume) were computed. 

In the event of algorithmic failure for a specific case, worst metric values were attributed. For bounded metrics (F1 and Mean Dice score) a value of 0 was given. For non-bounded error metrics (absolute element and absolute volume difference) an error of 100 000 was assigned as worst possible error.  

\begin{table}[hpb]
\small
\begin{tabular}{c|c|c|c|c}
\toprule
     & Detection Error & Detection Error& True Positive & True Positive \\
     \midrule
     & RefUnc $<=$ 0.5 & RefUnc $>$ 0.5 & RefUnc $<=$0.5 & RefUnc $>$ 0.5 \\
     PredUnc$<=$ 0.5 & FC & FC & TC & FC \\
     PredUnc $>$0.5 & TU & TU & FU & TC \\
     \bottomrule
\end{tabular}
\caption{Categorization for calculation of uncertainty measures; TU - Truly Uncertain; TC - Truly Certain; FU - Falsely Uncertain; FC - Falsely Certain}
\label{tab:unclac}
\end{table}

For Task 3 - Lacunes two metrics related to the estimation of uncertainty were further included. One was designed to tackle detection uncertainty and the other segmentation uncertainty. In terms of uncertainty validity, elements are considered as either truly certain (TC), truly uncertain (TU), falsely certain (FC) or falsely uncertain (FU) as per Table \ref{tab:unclac}.

The uncertainty was calculated as $(TU + TC)/(TU+TC +FC+FU)$.

The segmentation uncertainty was only assessed over true positive detected elements, assessing probabilistic uncertainty accuracy as $ \frac{\sum_{TP}(1-Unc) + \sum_{FN+FP}Unc}{TP+FN+FP}$ 

All metrics were computed per image and the distribution over all cases of the test set was used for the final ranking. For each task, ranking of the methods was performed following the method described for the Medical Image Decathlon challenge \citep{antonelli2021medical}. Pairwise comparisons were performed using the Mann-Whitney U-test for the Mean Dice over cases with F1 $>$ 0 and the Wilcoxon paired test for the other metrics due to their non-normal distribution. For each method, the number of times it was found significantly better (with a p-value $\leq$0.05 for significance) than another was used to rank the given metric. The final rank was obtained as the average across the ranks (lower being better). The robustness of the ranking was further assessed using the distribution of Kendall's tau correlation coefficient between ranking for all cases and the one obtained for 1000 bootstrap samples as described in \citep{wiesenfarth2021methods}.

To identify the best overall team, the ranks were averaged across all common metrics of all tasks for the teams that provided a solution to all three tasks.

\subsection{Additional analyses}
Further analyses were performed to inform on the following aspects: 1) clinical performance, 2) performance variability across datasets, 3) regional variability in performance (Task 1 - EPVS), 4) inter rater variability (Task 3 - Lacunes and part of Task 1 - EPVS),  and finally ensemble performance using either all methods (EnsembleAll) or the top 50\% (EnsembleTop).

\paragraph{Clinical performance}
For each task, the most clinically relevant metric was further defined and used to compare the different methods. For Task 1 - EPVS, to emphasize the notion of burden of EPVS, the correlation between predicted and reference volumes across the population of test cases was used. For Task 2 - Microbleeds and Task 3 - Lacunes where a binary statement of existence or absence is most clinically relevant, the balanced accuracy over cases considered as a whole-image classification task was chosen. 

\paragraph{Cross-dataset performance}
For each task, the performance of each method was additionally computed per dataset and then compared. The ranking was also computed per dataset to examine specific discrepancies between cohorts.

\paragraph{Regional performance}
To assess whether the performance of the proposed methods differed depending on the region for Task 1 - EPVS, the evaluation was run for each region (centrum semi-ovale, basal ganglia, hippocampus and mesencephalon) separately. For each method, pairwise comparison across regions was performed to assess whether a given method performed better on a given area. The overall ranking between methods was also computed per region.

\paragraph{Inter-rater variability}
For Task 1 - EPVS and Task 3 - Lacunes for which annotations by two raters were available, the evaluation was run considering alternatively each rater as the reference. While the overall absolute differences (volume and number of identified components) between the two raters are independent of the reference chosen (rater 1 or rater 2), changing the reference will affect F1 score and Mean Dice calculation due to differences in definition of true positives.


\paragraph{Ensemble performance}
Two ensemble solutions were created and evaluated. The average of all solutions (EnsembleAll) and the average of the predictions from the top 50\% in overall rank of the methods (EnsembleTop). EnsembleAll and EnsembleTop were compared to the individual methods for each task. The number of participating teams being 4 for Task 1 - EPVS, EnsembleTop in this case consists in the union of two best performing methods.

\section{Results}\label{sec2}
\subsection{Challenge submission and participating teams}
Over the period of the challenge, the data set has been requested for 353 downloads. Across the two validation periods, we received requests from 1 team at validation stage 1 and 4 teams at validation stage 2. The final submission of dockerized solutions and their documented description to be applied to the test sets was composed of 4 teams for Task 1 - EPVS, 9 teams for Task 2 - Microbleeds and 6 teams for Task 3 - Lacunes. Only 2 teams participated in all 3 tasks. Table \ref{tab:TaskPerTeam} summarizes in which task each team participated.

\begin{table}[h!]
    \centering
    \begin{tabular}{c|c|c|c}
      \textbf{Team Name}   & \makecell{\textbf{Task 1} \\ \textbf{EPVS}} & \makecell{\textbf{Task 2} \\ \textbf{Microbleeds}} & \makecell{\textbf{Task 3} \\ \textbf{Lacunes}}  \\
       BigrBrain  &  \checkmark & \checkmark & \checkmark \\
       Dawai & & \checkmark & \checkmark \\
       EMC\_N & & & \checkmark \\
       MixLacune & & & \checkmark \\
       MixMicrobleed & & \checkmark & \\
       MixMicrobleedNet & & \checkmark & \\
       Neurophet &\checkmark &  & \checkmark \\
       TeamTea & \checkmark& \checkmark& \checkmark \\
       Tfff & & \checkmark& \\
       TheGPU & \checkmark & \checkmark & \\
       ValdoNN  & & \checkmark & \\
       Zihao & & \checkmark & \\
    \end{tabular}
    \caption{Participation of the teams across the different tasks}
    \label{tab:TaskPerTeam}
\end{table}

Table \ref{tab:methods} reflects for each task and team the average time needed to evaluate one case, the GPU memory consumption, the docker details for memory requirements (CPU/GPU) and the methods' characteristics. The memory details are presented both as requested by the participants based on their training settings and as measured on a single case allowing for memory flooding. All methods using Stochastic Gradient Descent (SGD) as optimizer applied Nesterov Momentum with value of 0.99. Poly learning rate scheduling is defined as multiplying the learning rate by $\left(1-\frac{epoch}{epoch_{max}}\right)^{0.9}$. The following architectures were listed by the participating teams: 2D Unet \citep{ronneberger2015u}, 3D Unet \citep{cciccek20163d}, nnUnet \citep{isensee2021nnu}, MaskRCNN \cite{he2017mask}, Mask-RetinaNet \citep{farady2020mask}, ResNet \citep{he2016deep}. Beyond the well-known Dice \citep{milletari2016v} and binary cross-entropy losses, others such as focal loss \citep{lin2017focal} and blob loss \citep{kofler2022blob} were mentioned. Adam \citep{kingma2014adam}, SGD \citep{gardner1984learning} and Ranger21 \cite{wright2021ranger21} were the optimizers used.
\newpage
\newgeometry{left=0.5cm,bottom=1cm}
\begin{landscape}
\begin{table}
\captionsetup{font=tiny}
\renewcommand\cellalign{cc}
\setlength{\tabcolsep}{2pt}
  \centering
  \tiny
  \caption{Details of the methods of the participating teams for each task. Abbreviations: Aug. - Augmentation; BCE - Binary Cross Entropy; wBCE - weighted Binary Cross-Entropy; CSF - Cerebro spinal fluid; ES - Early Stopping; LR - Learning Rate; MAE - Mean Absolute Error; Mem - Memory; NM - Nesterov Momemtum (value 0.99); Norm. - Normalization; Optim. - Optimizer; PLRS - Poly learning rate schedule; Preproc. - Preprocessing; Pret. - Pretrained; Postproc. - Postprocessing;Req. - Requested; RF - Random Forest; SGD - Stochastic Gradient Descent; Val - Validation}
    \begin{tabular}{ccccccccccccccccccc}
          & \textbf{Team}     & \makecell{\textbf{Time}\\ \textbf{(min)}} & \makecell{\textbf{Mem} \\\textbf{(GB)}} & \textbf{Req} & \textbf{Method} & \textbf{Loss} & \textbf{Dim.}& \textbf{Input}& \textbf{Patches} & \textbf{Preproc.} & \textbf{Optim.} & \textbf{LR} & \makecell{\textbf{Stopping}\\ \textbf{criterion}} & \textbf{PostProc.} & \multicolumn{1}{c}{\textbf{Aug.}} & \textbf{Val\%} & \textbf{Framework} & \textbf{Pret} \\
    \midrule
    \multirow{4}[2]{*}{\begin{sideways}Task 1 - EPVS\end{sideways}} & BigrBrain & 0.87  & 1.9   & \makecell{32\\ 10} & UNet & Dice  &  2D    & All & 225 x 225 &    \makecell{Min-max Norm \\ Resampling \\ Cropping}   &    SGD NM  &  0.01 PLRS   &  \makecell{1000 \\ ES 10}      &      & \makecell{Rotation\\ Zooms\\ Shifts\\ Flips} & 20      &   Pytorch Ignite    &  \\
          & Neurophet & 1.92  & 1.7   &\makecell{10 \\ 20} & MaskRCNN & \makecell{BCE \\ Focal \\MAE} & 2.5D  & All &      & Norm &       &       &       &       &       &       &       &  \\
          & TeamTea & 1.38  & 3.7  & \makecell{10\\8} & nnUNet & Dice  & 2D  &All  & 256 x 224     & \makecell{Cropping\\ BF corr. \\ Z-score Norm. \\ Resampling} & SGD NM & 0.01 PLRS & 1000  &       &   \makecell{Zoom \\ Flip \\ Gaussian noise}    &   20    &       &  \\
          & TheGPU & 8.2   & NA & \makecell{10\\0}& RF    & NA    & 2D    &  T2 &    &  \makecell{Cropping \\ min-max Norm.}     &       &       &       &       &     33  &       &       &  \\
    \midrule
    \multirow{9}[2]{*}{\begin{sideways}Task 2 - Microbleeds\end{sideways}} & BigrBrain & 0.9   & 2.7   & \makecell{32\\10}& nnUNet  & Dice  & 2D  &All  & 512 x 512 &   \makecell{Min-max Norm \\ Resampling \\ Cropping}    &    SGD NM  & 0.01   PLRS   &   \makecell{1000\\ES 10}    &       & \makecell{Rotation\\ Zooms\\ Shifts\\  Flips} &  20     & Pytorch Ignite &  \\
          & Dawai & 11.2  & 42.2 &\makecell{128\\48}  & UNet & Blob  & 3D  & All  & 192 x 192 x 32     & Quantile Norm. & Ranger 21 &       &       &       & \makecell{ Flips\\  Gaussian Noise\\ Affine}  &       & MONAI & X \\
          & MixMicrobleed & 45.8  & 43.1 &\makecell{10\\10}  &\makecell{MaskRCNN\\UNet} &\makecell{Dice\\BCE\\MAE} & \makecell{2D \\ 2.5D}   & \makecell{64 x 64 \\ Whole}     & \makecell{Z score Norm. \\ Resampling} &   Adam    &   \makecell{0.000005 \\ 0.00005}    &  \makecell{15 \\ 50}     & X &   \makecell{Affine \\ Flips}    & 20      &       & X \\
          & MixMicrobleedNet & 1.4   & 3.2 &\makecell{10\\10}  & nnUNet & Dice  & 3D    &  All &    &       & SGD  NM    & 0.01   PLRS    &       &       &       &   0    &       &  \\
          & TeamTea & 2     & 3.3  & \makecell{10\\8} & UNet  & Dice  & 3D & All   & 96 x 192 x 128     &  \makecell{Z-score Norm\\ BF corr.\\ Resampling\\ Cropping} &  SGD NM    &  0.01  PLRS    & 1000        &       &   \makecell{Zoom\\Flips\\Noise}    & 30    & nnUNet &  \\
          & Tfff  & 1.5   & 6.5 & \makecell{10\\8}  & \makecell{ResNet \\UNet} & Dice  & 2D  & All  & 320x320 & min-max Norm. &       &       &       &       & \makecell{Flips \\ Rotation\\ Translation} &       & Pytorch Ignite &  \\
          & TheGPU & 5     & NA & \makecell{5\\0}& RF    &       & 2D    & T2* &    &       &       &       &       &       &       &       &       &  \\
          & ValdoNN & 0.6   & 2.3 & \makecell{10\\8}  & nnUNet &\makecell{Dice 
          \\BCE} & 2D    & All& 512 x 512     &  \makecell{Z-score Norm. \\ Resampling \\ Cropping}     &  SGD NM   & 0.01  PLRS     &  1000     &       &    \makecell{Rigid \\ Zoom \\ Gaussian noise}   &       &nnUNet &  \\
          & Zihao & 1.6   & 4.6 & \makecell{8\\6}   &\makecell{ UNet \\ FCN/AlexNet }& \makecell{wBCE \\ Dice} & 3D  & All  & \makecell{20x20x16 \\ 24x24x20}     & \makecell{z-score Norm \\ Resampling \\ Cropping} &  \makecell{SGD NM\\Adam}     &   0.01  PLRS  &  \makecell{150\\80\\100}     &       &    \makecell{Translation \\ Rotation \\ Flips}   &   20\% (5)    & Pytorch Ignite &  \\
    \midrule
    \multirow{6}[2]{*}{\begin{sideways}Task 3 - Lacunes\end{sideways}} & BigrBrain & 0.8   & 2  & \makecell{32\\10}   & UNet  & Dice  & 2D   & All & 384 x 320 &   \makecell{Min-max Norm \\ Resampling \\ Cropping}    &   SGD NM  &  0.01  PLRS   &  \makecell{1000\\ES 10}     &       & \makecell{Rotation\\ Zooms\\ Shifts\\ Flips}&       & Pytorch Ignite &  \\
          & Dawai & 11.9  & 42.2 & \makecell{128\\48}  & UNet  & Blob  & 3D  & All  & 192 x 192 x 32     &   \makecell{Quantile Norm. \\ Cropping}    & Ranger 21 &       &       &       &       &       & MONAI &  \\
          & EMC\_N & 32.1  & 42.7 & \makecell{24\\12} & UNet  & wBCE  & 3D  & All  &       & \makecell{Resampling \\ BF correction \\ Norm (wrt CSF)} &   Adam    & 0.0005 &    ES(20)   & X & \makecell{Rotation\\ Flips} & 10\% & Keras & X \\
          & MixLacune & 9.9   & 6.1 &\makecell{10\\10}  &\makecell{MaskRCNN  \\ UNet} & Dice  & 2D & All   &\makecell{64 x 64 \\32 x 32} & Z-score Norm.      &  Adam     &  \makecell{0.00005 \\ 0.0001}     &\makecell{20 \\ 30}       & X &   Flips    &       &       & X \\
          & Neurophet & 1.8   & 1.7 & \makecell{10\\20}  & MaskRetinaNet  & \makecell{BCE \\ Focal \\ MAE} & 2.5D  & All &      & Norm. & Adam & 0.0001 & \makecell{ReduceLR\\OnPlateau} &       &       &       &  &  \\
          & TeamTea & 2     & 3.3 & \makecell{10\\8}  & UNet  & Dice  & 3D & All   & 96 x 192 x 128     & \makecell{Z-score Norm\\ BF corr.\\ Resampling\\ Cropping}      &    SGD NM  &  0.01  PLRS   &   1000    &       &   \makecell{Zoom\\Flips\\Noise}    &       &    nnUNet   &  \\
    \bottomrule
    \end{tabular}%
  \label{tab:methods}%
\end{table}%

\end{landscape}
\restoregeometry
Among all the submissions, only one team (TheGPU) proposed an alternative to a deep learning solution. The majority of the proposed methods were trained as pure segmentation solutions and a few teams submitted a detection+segmentation solution based on Mask-RCNN \citep{he2017mask} or Mask Retina net \citep{farady2020mask}. Across all tasks, when a deep learning solution was proposed, the UNet architecture was the most common choice. For all three tasks, the time required to process a case and the GPU memory requirements varied greatly. For Task 2 - Microbleeds for instance duration ranged from less than 1 minute to 45.8 min and memory consumption of 2.4 to 43 GB (allowing for memory flooding).
In terms of the methodology for uncertainty assessments in Task 3 - Lacunes, the two teams submitting methods to all three tasks did not provide any uncertainty map. Among the 4 remaining teams, most used directly the probabilistic value of their output as measure of uncertainty while mixLacune defined an uncertainty zone at the border of their detected lacunes. 

For all teams, key characteristics of the proposed methods are summarized in table \ref{tab:methods}. Additional details can be found for each team on the OpenReview repository \url{https://openreview.net/group?id=MICCAI.org/2021/Challenge/VALDO}. 

\subsection{Metric values}
For each task the detection and the segmentation are reported across all teams.
\paragraph{Task 1 - Enlarged Perivascular Spaces (EPVS)}
The summary statistics for each team and each metric are reported in Table \ref{tab:results_pvs}.

\begin{table}[htbp]
  \centering
  \tiny
  \caption{Metrics results for Task 1 - EPVS presented as Median [1st quartile - 3rd quartile] for all metrics. AED - Absolute Element Difference; AVD (in mm3) - Absolute Volume Difference. In bold the significantly best performance across the different teams (excluding the ensemble solutions) and in italic when there is no significant difference compared to the second best.}
    \begin{tabular}{lllll}
    \toprule
          & \multicolumn{2}{c}{Detection} & \multicolumn{2}{c}{Segmentation} \\
          & \multicolumn{1}{c}{F1} & \multicolumn{1}{c}{AED} & \multicolumn{1}{c}{Mean Dice} & \multicolumn{1}{c}{AVD} \\
    \midrule
    Bigrbrain & \multicolumn{1}{c}{35.81 [28.14 ; 40.42]} & \multicolumn{1}{c}{\textit{14.50 [6.00 ; 34.50]}} & \multicolumn{1}{c}{61.09 [55.40 ; 66.57]} & \multicolumn{1}{c}{45.30 [16.12 ; 89.12]} \\
    Neurophet & \multicolumn{1}{c}{0.00 [0.00 ; 3.34]} & \multicolumn{1}{c}{29.00 [13.00 ; 47.00]} & \multicolumn{1}{c}{28.23 [23.27 ; 29.76]} & \multicolumn{1}{c}{390.15 [250.72 ; 636.58]} \\
    TeamTea & \multicolumn{1}{c}{17.12 [6.79 ; 25.90]} & \multicolumn{1}{c}{41.00 [24.25 ; 69.25]} & \multicolumn{1}{c}{55.07 [46.25 ; 64.23]} & \multicolumn{1}{c}{106.05 [73.00 ; 175.86]} \\
    TheGPU & \multicolumn{1}{c}{\textbf{38.92 [28.87 ; 49.44]}} & \multicolumn{1}{c}{16.00 [9.00 ; 35.75]} & \multicolumn{1}{c}{\textbf{72.38 [64.97 ; 77.12]}} & \multicolumn{1}{c}{\textit{45.20 [23.79 ; 82.21]}} \\
    \midrule
    EnsembleAll & 38.62 [28.1 ; 44.82] & 24.00 [12.00 ; 46.00] & 64.33 [59.14 ; 68.40] & 96.15 [63.67 ; 151.69] \\
    EnsembleTop & 38.86 [31.19 ; 45.13] & 29.00 [15.25 ; 50.25] & 67.38 [58.24 ; 72.23] & 36.10 [20.15 ; 66.33] \\
    \bottomrule
    \end{tabular}%
  \label{tab:results_pvs}%
\end{table}%

\begin{figure}
    \centering
    \includegraphics[width=0.95\textwidth]{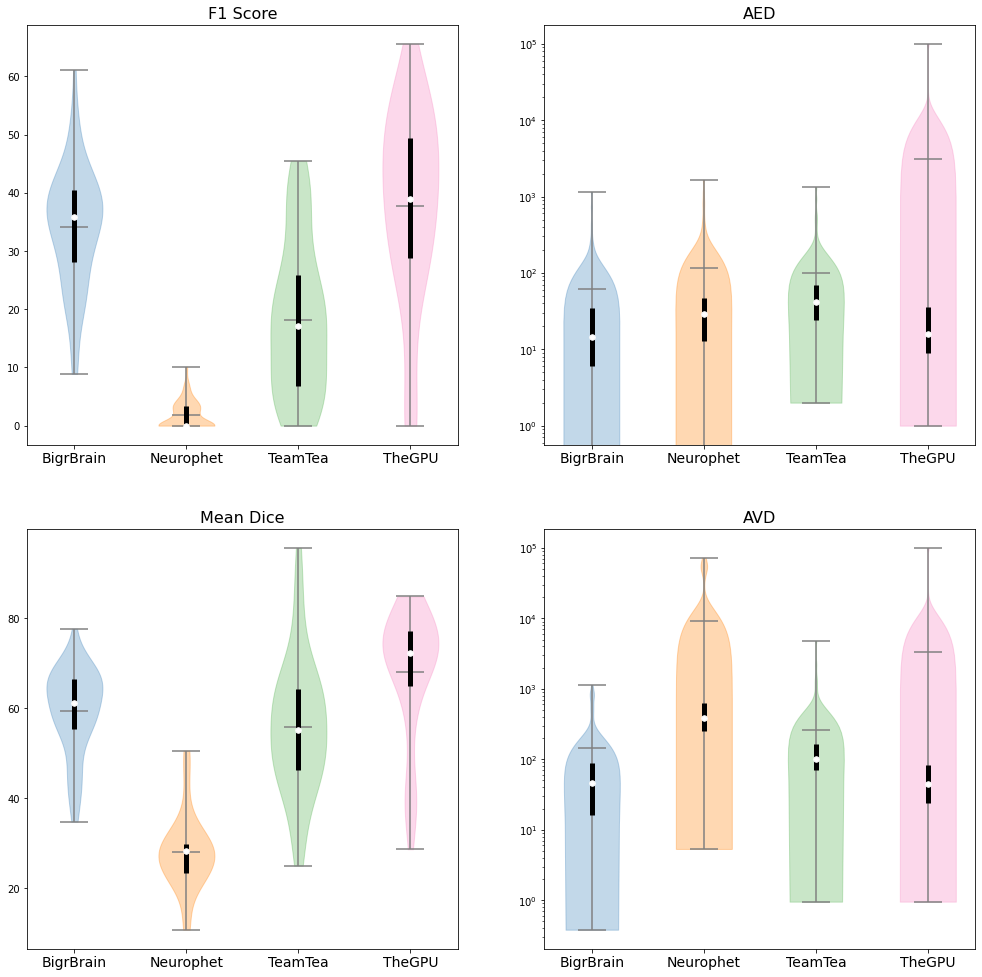}
    \caption{Distribution of metrics values across the different teams for detection metrics (top row) and segmentation metrics (bottom row) for Task 1 - EPVS}
    \label{fig:results_pvs}
\end{figure}

Figure \ref{fig:results_pvs} presents the distribution of metrics values for detection (top row) and segmentation metrics (bottom row) for Task 1 - EPVS. 

\paragraph{Task 2 - Microbleeds}
\begin{table}[htbp]
  \centering
  \tiny
    \begin{tabular}{lcccc}
    \toprule
          & \multicolumn{2}{c}{Detection} & \multicolumn{2}{c}{Segmentation} \\
          & F1    & AED    & Mean Dice & AVD \\
    \midrule
    Bigrbrain & 16.67 [0.00 ; 36.10] & 9.00 [5.00 ; 16.00] & 81.17 [71.86 ; 89.69] & 52.47 [15.45 ; 171.98] \\
    Dawai & 0.00 [0.00 ; 40.00] & 1.00 [1.00 ; 3.00] & 68.35 [52.99 ; 77.71] & 12.40 [6.29 ; 33.05] \\
    MixMicrobleed & 0.00 [0.00 ; 0.00] & 1e5 [499.5 ; 1e5] & 64.36 [55.79 ; 68.58] & 1e5 [4728 ; 1e5] \\
    MixMicrobleedNet & \textbf{68.42 [36.67 ; 100.00]} & \textbf{1.00 [0.00 ; 1.00]} & \textbf{84.01 [79.48 ; 87.62]} & \textbf{8.77 [2.48 ; 24.30]} \\
    TeamTea & 66.67 [0.00 ; 100.00] & \textbf{1.00 [0.00 ; 1.00]} & 82.57 [74.65 ; 87.50] & 11.30 [1.81 ; 25.39] \\
    Tfff  & 40.00 [18.18 ; 66.67] & 3.00 [1.00 ; 6.00] & 77.65 [62.43 ; 89.13] & 15.27 [4.33 ; 49.33] \\
    TheGPU & 0.00 [0.00 ; 0.00] & 4.00 [1.00 ; 10.00] & 49.46 [36.89 ; 78.14] & 602.89 [159 ; 1842.02] \\
    ValdoNN & 50.00 [0.00 ; 68.15] & 1.00 [1.00 ; 2.00] & 80.00 [66.67 ; 87.68] & 12.00 [3.14 ; 24.91] \\
    Zihao & 66.67 [20.83 ; 100.00] & 1.00 [0.00 ; 2.00] & 80.00 [73.34 ; 88.04] & 9.61 [3.20 ; 21.51] \\
    \midrule
    EnsembleAll & 66.67 [0.00 ; 100.00] & 1.00 [0.00 ; 1.00] & 81.22 [71.35 ; 87.27] & 12.87 [4.93 ; 27.26] \\
     EnsembleTop & 75.68 [38.18 ; 100.00] & 1.00 [0.00 ; 1.00] & 77.90 [29.91 ; 87.23] & 11.25 [2.81 ; 21.82]\\
    \bottomrule
    \end{tabular}%
    \caption{Metrics results for Task 2 - Microbleeds presented as Median [1st quartile; 3rd quartile] for each metric. AED - Absolute Element difference; AVD - Absolute volume difference (in mm3). In bold, the significantly best performance per metric across teams (excluding the ensemble solutions)}.
  \label{tab:results_cmb}%
\end{table}%

\begin{figure}
    \centering
    \includegraphics[width=0.95\textwidth]{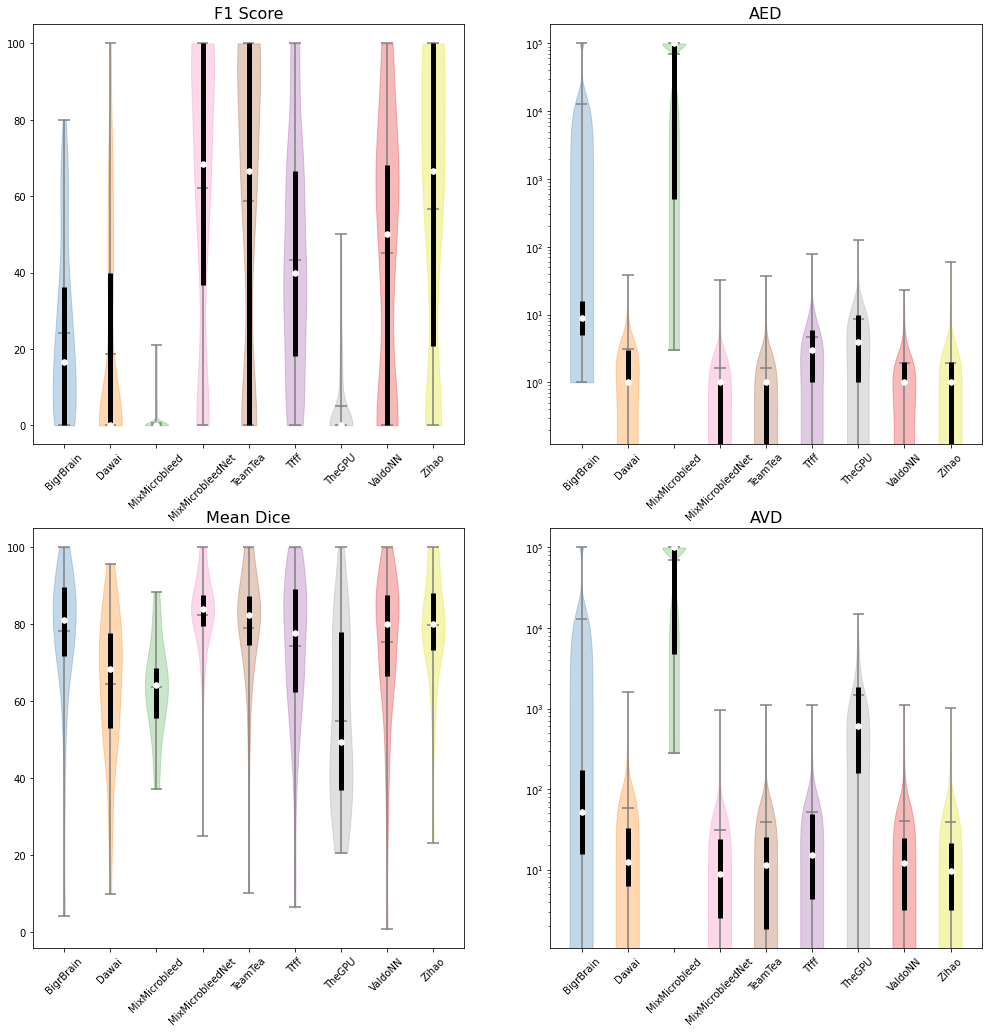}
    \caption{Distribution of metrics values across the different teams for detection metrics (top row) and segmentation metrics (bottom row) for Task 2 - Microbleeds; AED - Absolute Element Difference; AVD - Absolute Volume Difference}
    \label{fig:results_cmb}
\end{figure}

Figure \ref{fig:results_cmb} presents the distribution of metrics values for detection (top row) and segmentation metrics (bottom row) for Task 2 - Microbleeds with Table \ref{tab:results_cmb} presenting the metrics values across all teams. 

\paragraph{Task 3 - Lacunes}

Table \ref{tab:results_lac} presents the results obtained for Task 3 - Lacunes.

\begin{table}[htbp]
  \centering
  \tiny
    \begin{tabular}{lllll}
    \toprule
          & \multicolumn{2}{c}{\textbf{Detection}} & \multicolumn{2}{c}{\textbf{Segmentation}} \\
          & \textbf{F1} & \textbf{AED} & \textbf{Mean Dice} & \textbf{AVD} \\
    \midrule
   BigrBrain & 7.69 [5.06 ; 16.49] & 27.50 [20.25 ; 33] & 40.84 [27.02 ; 50.27] & 123.93 [79.49 ; 182.64] \\
    Dawai & 15.38 [0.00 ; 25.00] & 6.00 [3.00 ; 10.00] & 40.09 [26.20 ; 45.31] & 78.93 [26.94 ; 209.24] \\
    EMC\_N & 3.92 [0.00 ; 54.55] & 2.00 [1.00 ; 4.75] & 20.49 [12.21 ; 34.08] & 125.60 [45.08 ; 375.96] \\
    MixLacune & 6.25 [0.00 ; 12.00] & 22.00 [13.50 ; 26.00] & 16.85 [10.31 ; 27.59] & 33.95 [16.88 ; 107.69] \\
    Neurophet & 4.55 [0.00 ; 10.53] & 20.00 [11.50 ; 34.00] & 8.82 [3.73 ; 15.33] & 471.40 [244.16 ; 891.16] \\
    TeamTea & \textit{28.57 [0.00 ; 57.14]} & \textbf{1.00 [0.00 ; 2.00]} & \textbf{45.75 [36.74 ; 56.17]} & \textbf{14.88 [0.00 ; 40.29]}\\
    \midrule
    EnsembleAll & 28.57 [0.00 ; 60.87] & 1.00 [0.00 ; 2.00] & 37.98 [22.13 ; 44.55] & 13.05 [0.07 ; 61.03] \\
          EnsembleTop & 30.77 [0.00 ; 66.67] & 1.00 [0.00 ; 2.00] & 38.17 [25.48 ; 45.26] & 9.68 [1.05 ; 63.28] \\
    \bottomrule
    \end{tabular}%
    \caption{Metrics results for Task 3 - Lacunes presented as median [1st quartile ; 3rd quartile]. AED - Absolute Element difference; AVD - Absolute volume difference (mm3). Bold font indicates best performance across the teams (excluding ensemble solutions) when significantly better than all others. Italic font indicates best performance when not significantly better than the second ranking}
  \label{tab:results_lac}%
\end{table}%

while Table \ref{tab:lac_unc} shows the metrics for the uncertainty component of the task excluding BigrBrain and TeamTea who did not provide an uncertainty map.

\begin{table}[htbp]
  \centering
  \caption{Metrics related to uncertainty for Task 3 - Lacunes presented as median [1st quartile - 3rd quartile. AED - Absolute Element difference; AVD - Absolute volume difference (in mm3).}
    \begin{tabular}{lll} 
    \toprule
          & \textbf{Detection Unc} & \textbf{Segmentation Unc} \\
    \midrule
   \textbf{Dawai} & 0.00 [0.00 ; 25.00] & 63.65 [0.00 ; 87.73] \\
    \textbf{EMC\_N} & 100.00 [86.81 ; 100.00] & 0.00 [0.00 ; 67.94] \\
    \textbf{MixLacune} & 0.00 [0.00 ; 3.57] & 4.76 [0.00 ; 24.39] \\
    \textbf{Neurophet} & 0.00 [0.00 ; 6.82] & 0.00 [0.00 ; 23.18] \\
    \bottomrule
    \end{tabular}%
  \label{tab:lac_unc}%
\end{table}%

\begin{figure}
    \centering
    \includegraphics[width=0.95\textwidth]{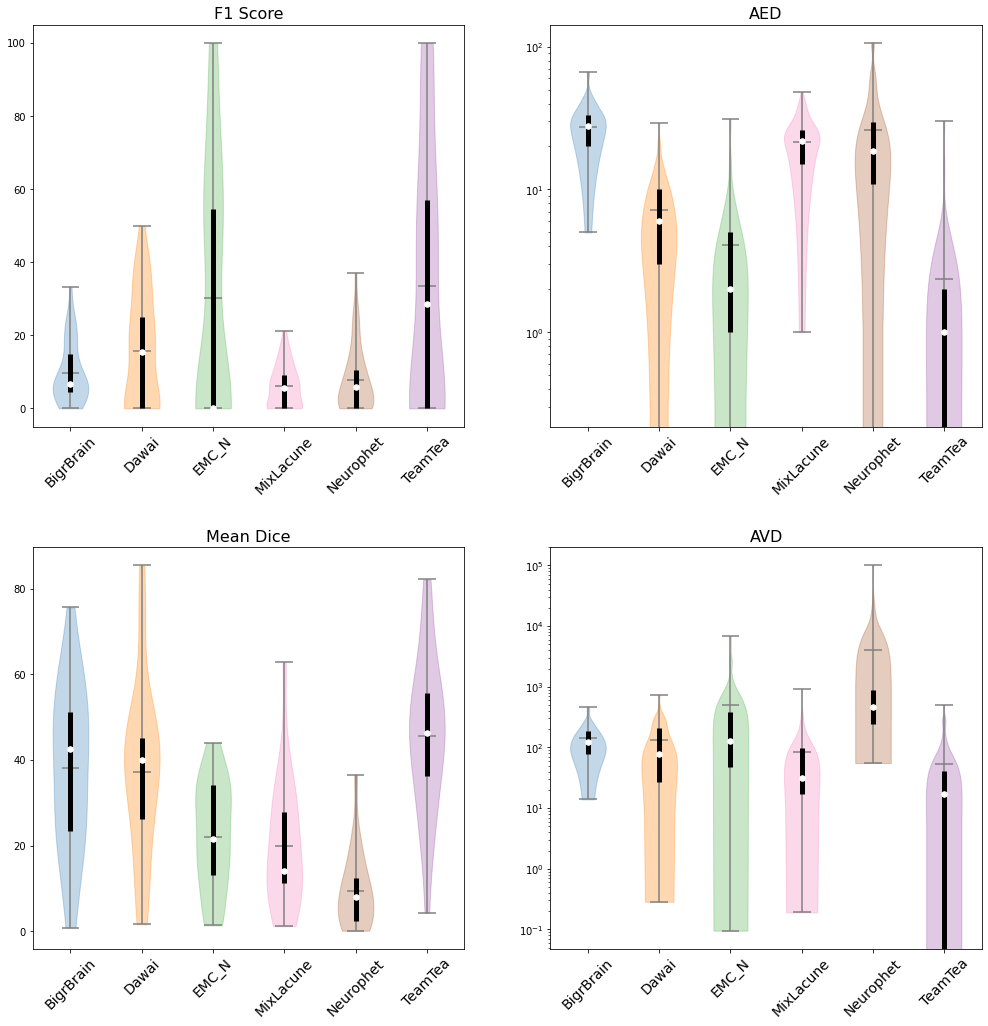}
    \caption{Distribution of metric values across the different teams for detection metrics (top row) and segmentation metrics (bottom row) for Task 3 - Lacunes}
    \label{fig:results_lac}
\end{figure}

Figure \ref{fig:results_lac} presents the distribution of metrics values for detection (top row) and segmentation metrics (bottom row) for Task 3 - Lacunes. 

\begin{figure}
    \centering
    \includegraphics[width=0.95\textwidth]{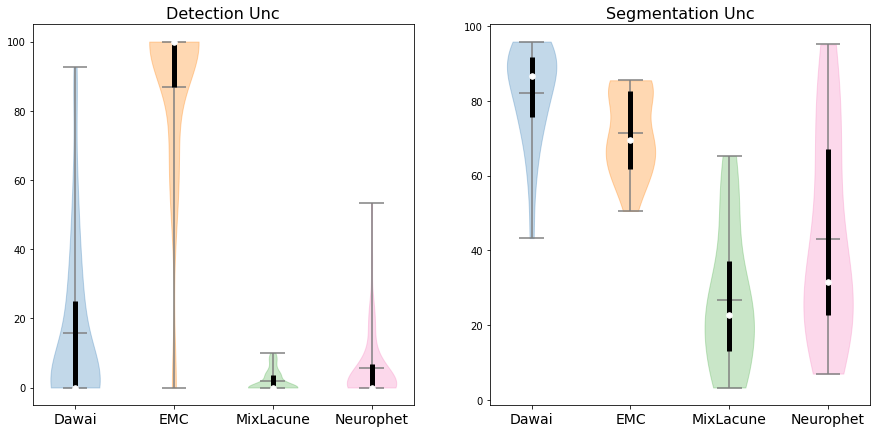}
    \caption{Distribution of metric values across the different teams for the assessment of uncertainty for Task 3 - Lacunes}
    \label{fig:results_unc}
\end{figure}

Figure \ref{fig:results_unc} shows the distribution of metrics values for the assessment of uncertainty applied for Task 3 - Lacunes.
\subsection{Rankings}
         
   

Table \ref{tab:ranking_all} presents the overall ranking, according to the number of tasks undergone and for each individual task when relevant.

\begin{table}[htbp]
  \centering
  \tiny
  \setlength{\tabcolsep}{3pt}
  \caption{Ranking across all tasks grouped by number of tasks to which each team participated. Across all metrics, D refers to detection and S to segmentation, R to relative, A to absolute and U to uncertainty. DR refers to F1 score, DA to Absolute element difference, SR to Mean Dice, SA to absolute volume difference, DU to detection uncertainty and SU to segmentation uncertainty. Tot is the overall rank for a given task}
    \begin{tabular}{l|ccccc|ccccc|ccccccc}
\cmidrule{2-18} 
& \multicolumn{5}{c|}{\textbf{Task 1 - EPVS}} & \multicolumn{5}{c|}{\textbf{Task 2 - Microbleeds}} & \multicolumn{7}{c|}{\textbf{Task 3 - Lacunes}}  \\
    \midrule
    \textbf{Team} & \textbf{DR} & \textbf{DA} & \textbf{SR} & \textbf{SA} & \textbf{Tot}& \textbf{DR} & \textbf{DA} & \textbf{SR} & \textbf{SA} & \textbf{Tot}& \textbf{DR} & \textbf{DA} & \textbf{SR} & \textbf{SA} & \textbf{DU}& \textbf{SU}&\textbf{Tot} \\
    TeamTea & 3   & 3.5     & 3     & 2.5 & 3  & 1.5   & 2.5   & 2.5   & 3  & 2   & 1.5     & 1   & 1     & 1   &   & & \textbf{2} \\
    BigrBrain & 2     & 1.5    & 2     & 1  & 2   & 6     & 8     & 3.5   & 7    & 6  & 4   & 5.5     & 2.5   & 4.5  &   && \textbf{6} \\
    \midrule
    Dawai &  &  &  &      &  & 5     & 7     & 7.5   & 5.5 & 7  & 3     & 3     & 2.5   & 3  &2 &1 &  \textbf{1} \\
    TheGPU & 1     & 1.5     & 1     & 2.5   & 1 & 7     & 8     & 9     & 8  &  8 &  &  &  &  & & &  \\
    Neurophet & 4   & 3.5     & 4     & 4  & 4  &  &  &  & & & 5.5   & 5.5   & 6     & 6  & 3  &3.5 & \textbf{5} \\
    \midrule
    MixMicrobleedNet &  &  &  &    &   & 1.5   & 1     & 1     & 1  & 1   &  &  &  &  & & & \\
    Zihao Team &  &  &  &      & & 3     & 2.5   & 2.5   & 3  & 3  &  &  &  &  & & &  \\
    ValdoNN &  &  &  & &  & 4     & 4.5   & 5     & 3   & 4 &  &  &  &  & & &  \\
    Tfff  &  &  &  &  & & 6     & 4.5   & 5     & 5.5  &5 &  &  &  &  && &  \\
    MixMicrobleed &  &  &  &  & & 9     & 9     & 7.5   & 9  & 9  &  &  &  &  && & \\
    \midrule
    EMC\_N &  &  &  &  &  &  &  & && & 1.5   & 2   & 4.5     & 4.5   &  1&2 & \textbf{2}\\
    MixLacune &  &  &  &  &  &&&  &  &  & 5.5   & 4   & 4.5     & 2  & 4  &3.5 & \textbf{4}\\
    \bottomrule
    \end{tabular}%
    
  \label{tab:ranking_all}%
\end{table}%

Table \ref{tab:kt} reflects the distribution of Kendall's Tau coefficient when assessing the robustness of the ranking for each metric using 1000 bootstrap samples.

\begin{table}[htbp]
  \centering
  \tiny
  \caption{Distribution characteristics (mean and standard deviation) Kendall's Tau correlation coefficient in \% between final ranking and bootstrap samples (1000 samples). Across all metrics, D refers to detection and S to segmentation, R to relative, A to absolute and U to uncertainty. DR refers to F1 score, DA to Absolute element difference, SR to Mean Dice, SA to absolute volume difference, DU to detection uncertainty and SU to segmentation uncertainty.}
    \begin{tabular}{lcccccc}
    \toprule
          & \textbf{DR} & \textbf{DA} & \textbf{SR} & \textbf{SA} & \textbf{DU} & \textbf{SU} \\
    \midrule
    \textbf{Task 1 - EPVS} & 96.13 (4.33) & 93.55 (7.39) & 97.87 (4.45) & 97.33 (4.02) &       &  \\
    \textbf{Task 2 - Microbleeds} & 98.11 (1.81) & 98.36 (1.70) & 98.19 (2.38) & 87.08 (6.62) &       &  \\
    \textbf{Task 3 - Lacunes} & 95.88 (6.57) & 97.46 (3.98) & 94.68 (3.26) & 93.19 (5.02) & 99.85 (1.13) & 95.82 (8.37) \\
    \bottomrule
    \end{tabular}%
  \label{tab:kt}%
\end{table}%

\subsection{Additional analyses}
\subsubsection{Clinical relevant markers}
\begin{figure}
    \centering
    \includegraphics[width=0.95\textwidth]{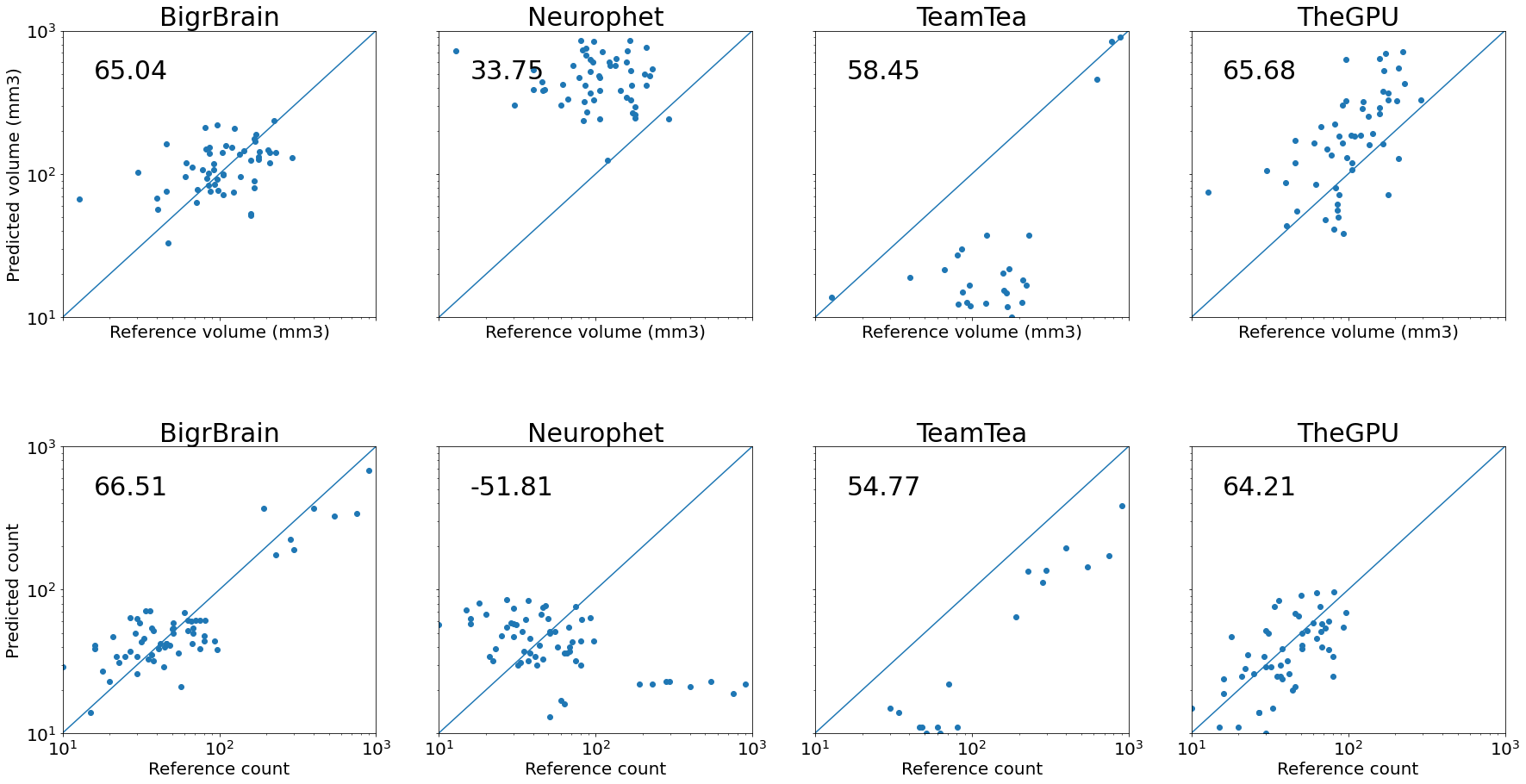}
    \caption{Association between reference and predicted PVS burden across the participating teams for volume (top row) and count (bottom row). The Spearman rho (\%) is indicated on each plot.}
    \label{fig:clinical_pvs}
\end{figure}

\paragraph{Task 1 - EPVS}
For Task 1, since the burden of PVS is currently clinically considered the most valuable insight, the Spearman correlation coefficient between predicted and reference burden across all test cases was calculated for overall volume and element count and is presented in Figure \ref{fig:clinical_pvs} along with the log-transformed  relationship between reference and predicted burden in terms of volume (top) and count (bottom).

\begin{figure}[htbp]
    \centering
    \includegraphics[width=0.95\textwidth]{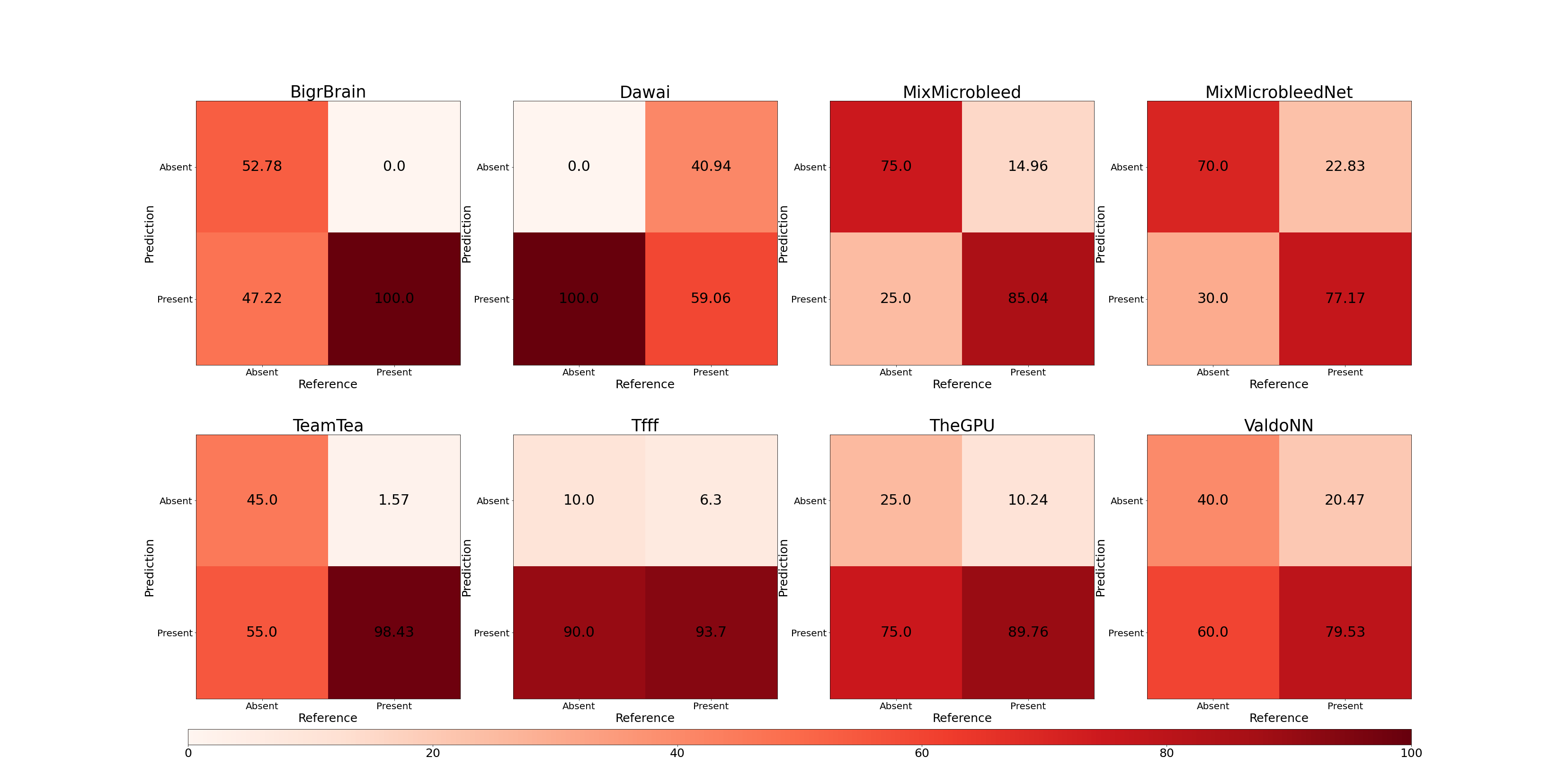}
    \caption{Confusion matrix regarding the classification of an image as containing at least one microbleed based on obtained prediction images.}
    \label{fig:clinical_cmb}
\end{figure}

\paragraph{Task 2 - Microbleeds}
For cerebral microbleeds, classifying the absence or presence of any microbleeds was deemed clinically the most relevant assessment. 
Balanced accuracy over the test set varied from 29.5\% for team Dawai to 87.3\% for team MixMicrobleed. Figure \ref{fig:clinical_cmb} presents the confusion matrices for each of the teams.

\begin{figure}[htbp]
    \centering
    \includegraphics[width=0.95\textwidth]{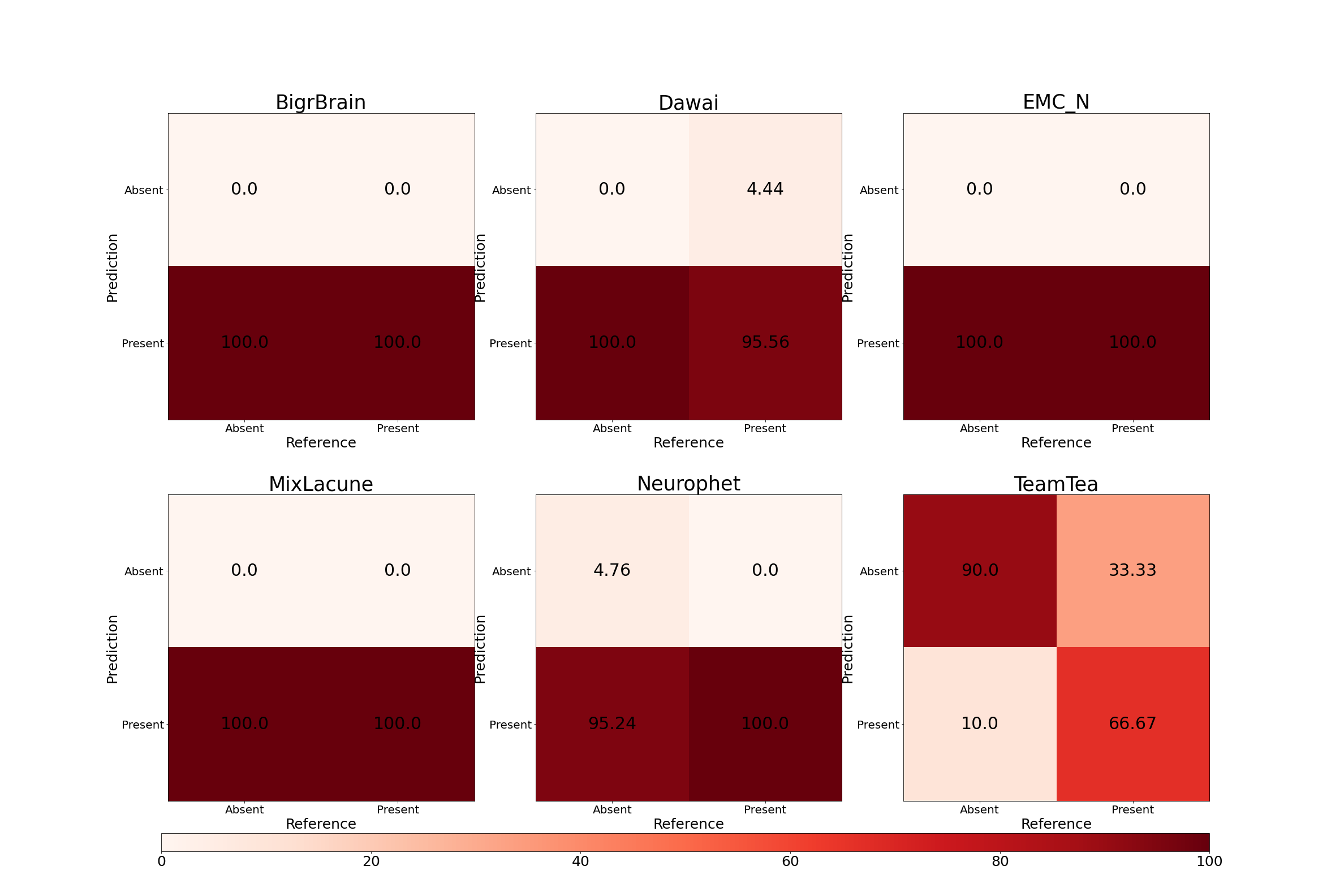}
    \caption{Confusion matrix regarding the classification of an image as containing at least one lacune based on obtained prediction images.}
    \label{fig:clinical_lac}
\end{figure}

\paragraph{Task 3 - Lacunes}
Similarly, Figure \ref{fig:clinical_lac} shows the confusion matrix for correctly identifying cases that have at least one lacune. For the 6 participating teams, balanced accuracy was close to 0.5 for almost all teams as they predicted the presence of at least one lacune in almost all cases. Only TeamTea was able to recognize cases without lacunes, with 78.3\% balanced accuracy. 

\subsubsection{Cross-dataset variability}
\begin{figure}
    \centering
    \begin{tabular}{c}
    
     \includegraphics[width=0.7\textwidth]{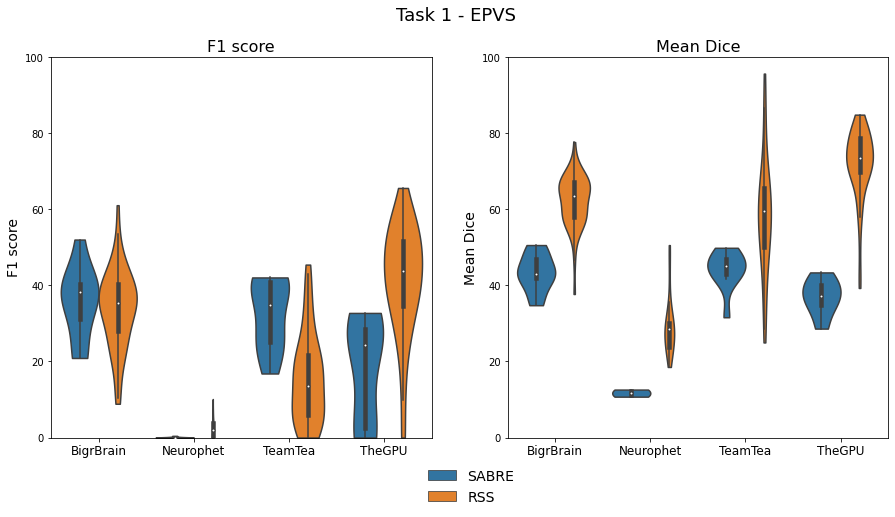}    \\ \includegraphics[width=0.7\textwidth]{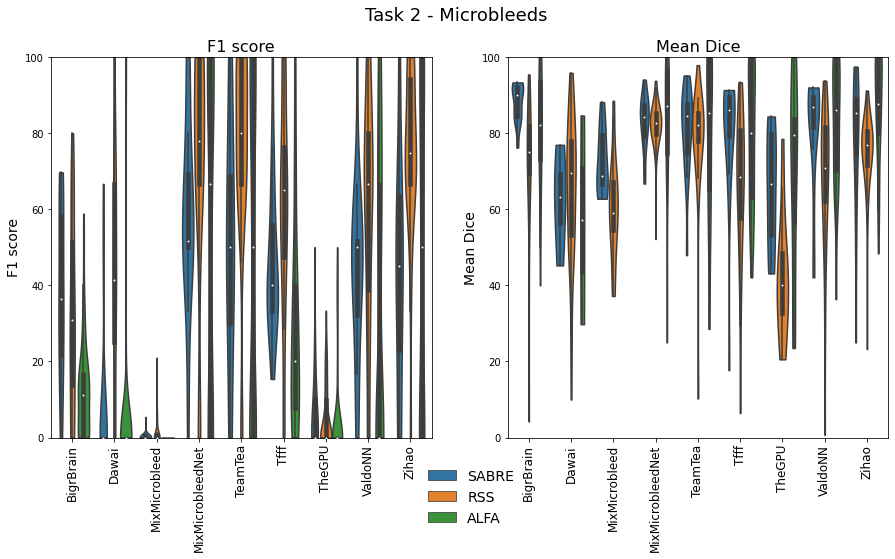}  \\ \includegraphics[width=0.7\textwidth]{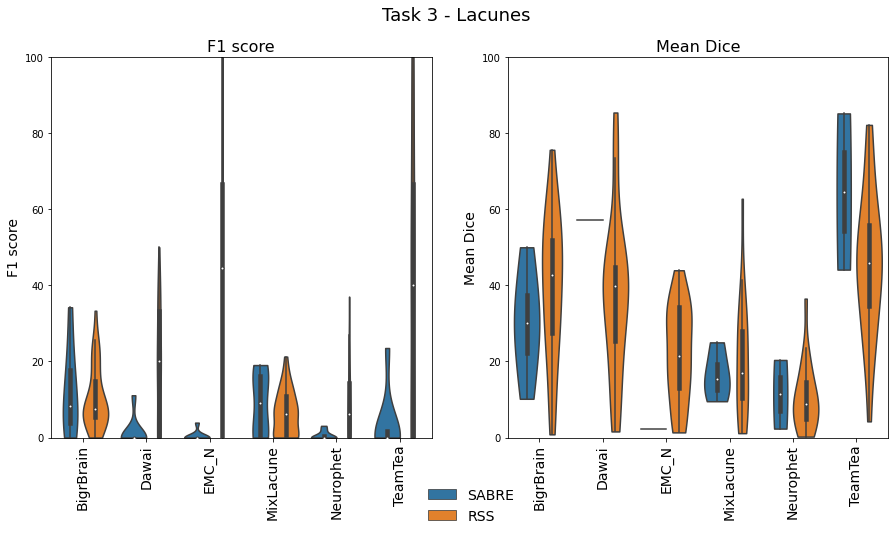}  \\
    \end{tabular}
    
    \caption{Distribution of results for F1 (left column) and Mean Dice (right column) across different datasets for the three tasks (each row represents one task). }
    \label{fig:dataset_all}
\end{figure}

Performance varied greatly across datasets, being systematically overall better on RSS dataset than others (SABRE or ALFA). 
For all three tasks, Figure \ref{fig:dataset_all} presents the variation of F1 and Mean Dice score across datasets for all teams and Table \ref{tab:datasets_all_metrics} presents median and interquartile range for all tasks across datasets for F1 score and Mean Dice.


\begin{table}[htbp]
  \centering
  \tiny
  \setlength{\tabcolsep}{3pt}
  \caption{F1 score and Mean Dice presented as median [1st quartile; 3rd quartile] across the different datasets for all three tasks}
    \begin{tabular}{cccccc|ccc}
    \toprule
          &       & \multicolumn{3}{c|}{\textbf{F1 Score}} & \multicolumn{3}{c}{\textbf{Mean Dice}} &  \\
          &       &\textbf{ALFA} & \textbf{RSS} & \textbf{SABRE} & \textbf{ALFA} & \textbf{RSS} & \textbf{SABRE} &  \\
    \midrule
    \multirow{4}[2]{*}{\begin{sideways}\textbf{Task 1}\end{sideways}} & BigrBrain &       & 35.40 [27.89 ; 40.36] & 38.20 [31.19 ; 40.25] &       & 63.56 [58.05 ; 67.29] & 43.10 [41.94 ; 46.92] &  \\
          & Neurophet &       & 1.95 [0.00 ; 3.71] & 0.00 [0.00 ; 0.00] &       & 28.6 [23.90 ; 30.17] & NA    &  \\
          & TeamTtea &       & 13.50 [6.02 ; 21.76] & 34.80 [25.10 ; 40.93] &       & 59.66 [50.08 ; 65.55] & 45.05 [42.88 ; 46.91] &  \\
          & TheGPU &       & 43.71 [34.55 ; 51.67] & 24.32 [2.61 ; 28.42] &       & 73.41 [69.68 ; 78.6] & 34.96 [7.16 ; 39.25] &  \\
    \midrule
    \multirow{9}[2]{*}{\begin{sideways}\textbf{Task 2}\end{sideways}} & BigrBrain & 11.11 [0.00 ; 16.67] & 30.77 [13.81 ; 51.47] & 36.36 [21.81 ; 58.24] & 82.21 [73.3 ; 93.32] & 75.07 [69.41 ; 81.84] & 90.13 [84.63 ; 92.19] &  \\
          & Dawai & 0.00 [0.00 ; 0.00] & 41.43 [25.00 ; 66.67] & 0.00 [0.00 ; 0.00]  & 57.20 [43.49 ; 70.91] & 69.47 [53.16 ; 77.82] & 63.33 [56.31 ; 69.23] &  \\
          & Mixmicrobleed & 0.00 [0.00 ; 0.00] & 0.00 [0.00 ; 0.75] & 0.00 [0.00 ; 0.42]  & 0.00 [0.00 ; 0.00] & 58.90 [54.56 ; 67.04] & 68.62 [66.67 ; 79.41] &  \\
          & MixmicrobleedNet & 66.67 [0.00 ; 100] & 77.81 [66.67 ; 100.00] & 51.67 [50.00 ; 69.23] & 87.18 [74.71 ; 96.67] & 82.79 [79.82 ; 85.20] & 84.21 [79.35 ; 87.39] &  \\
          & TeamTea & 50.00 [0.00 ; 100.00] & 80.00 [66.67 ; 100.00] & 50.00 [30.22 ; 68.75] & 85.16 [65.38 ; 100.00] & 82.08 [77.83 ; 85.24] & 84.62 [74.65 ; 87.66] &  \\
          & Tfff  & 20.00 [7.68 ; 40.00] & 65.15 [47.50 ; 76.41] & 40.00 [33.33 ; 55.91] & 80.00 [63.19 ; 99.46] & 68.58 [57.64 ; 80.72] & 86.19 [79.14 ; 89.46] &  \\
          & TheGPU & 0.00 [0.00 ; 0.00] & 0.00 [0.00 ; 9.95] & 0.00 [0.00 ; 10.01] & 79.43 [56.53 ; 83.76] & 40.00 [32.63 ; 48.54] & 66.67 [53.28 ; 79.70] &  \\
          & ValdoNN & 0.00 [0.00 ; 66.67] & 66.67 [38.82 ; 80.00] & 50.00 [32.14 ; 51.56] & 86.06 [70.24 ; 100.00] & 70.91 [62.08 ; 81.48] & 87.00 [81.51 ; 89.58] &  \\
          & Zihao & 50.00 [0.00 ; 100.00] & 74.81 [66.67 ; 94.23] & 45.00 [22.92 ; 63.54] & 87.71 [80.00 ; 100.00] & 76.98 [71.69 ; 80.46] & 85.42 [74.53 ; 88.85] &  \\
\midrule
\multirow{6}[2]{*}{\begin{sideways}\textbf{Task 3}\end{sideways}} & BigrBrain &       & 7.41 [5.48 ; 14.91] & 8.39 [3.80 ; 17.75] &       & 42.81 [27.39 ; 51.80] & 30.17 [22.30 ; 37.37] &  \\
          & Dawai &       & 20.00 [0.00 ; 33.33] & 0.00 [0.00 ; 0.00] &       & 39.88 [25.38 ; 44.82] & 57.14 [57.14 ; 57.14] &  \\
          & EMC\_N &       & 44.44 [0.00 ; 66.67] & 0.00 [0.00 ; 0.00] &       & 21.42 [13.09 ; 34.24] & 2.20 [2.20 ; 2.20] &  \\
          & MixLacune &       & 6.25 [0.00 ; 10.81] & 9.09 [0.00 ; 16.16] &       & 16.85 [10.31 ; 28.04] & 15.45 [12.51 ; 19.3] &  \\
          & Neurophet &       & 6.25 [0.00 ; 14.29] & 0.00 [0.00 ; 0.46] &       & 8.82 [4.98 ; 14.65] & 11.37 [6.86 ; 15.88] &  \\
          & TeamTea &       & 40.00 [0.00 ; 66.67] & 0.00 [0.00 ; 1.61] &       & 45.75 [34.58 ; 55.75] & 64.66 [54.40 ; 74.92] &  \\
\bottomrule   \end{tabular}%
  \label{tab:datasets_all_metrics}%
\end{table}%

Ranking varied also slightly across datasets as indicated in Table \ref{tab:dataset_ranking}.

\begin{table}[htbp]
  \centering
  \caption{Ranking calculated for each dataset separately}
    \begin{tabular}{clrcc}
\cmidrule{3-5}          &       & \multicolumn{1}{c}{\textbf{ALFA}} & \textbf{RSS} & \textbf{SABRE} \\
    \midrule
    \multirow{4}[2]{*}{\begin{sideways}\textbf{Task 1 - EPVS}\end{sideways}} & \textbf{BigrBrain} &       & 2     & 1 \\
          & \textbf{Neurophet} &       & 4     & 4 \\
          & \textbf{TeamTea} &       & 3     & 2 \\
          & \textbf{TheGPU} &       & 1     & 3 \\
    \midrule
    \multirow{9}[2]{*}{\begin{sideways}\textbf{Task 2 - Microbleeds}\end{sideways}} & \textbf{BigrBrain} & \multicolumn{1}{c}{7} & 7     & 6 \\
          & \textbf{Dawai} & \multicolumn{1}{c}{6} & 6     & 7 \\
          & \textbf{MixMicrobleed} & \multicolumn{1}{c}{9} & 9     & 9 \\
          & \textbf{MixMicrobleedNet} & \multicolumn{1}{c}{2} & 1     & 2 \\
          & \textbf{TeamTea} & \multicolumn{1}{c}{3} & 2     & 1 \\
          & \textbf{Tfff} & \multicolumn{1}{c}{5} & 5     & 5 \\
          & \textbf{TheGPU} & \multicolumn{1}{c}{8} & 8     & 8 \\
          & \textbf{ValdoNN} & \multicolumn{1}{c}{4} & 4     & 3 \\
          & \textbf{Zihao} & \multicolumn{1}{c}{1} & 3     & 4 \\
    \midrule
    \multirow{6}[2]{*}{\begin{sideways}\textbf{Task 3 - Lacunes}\end{sideways}} & \textbf{BigrBrain} &       & 5     & 5 \\
          & \textbf{Dawai} &       & 2.5   & 1 \\
          & \textbf{EMC\_N} &       & 1     & 3 \\
          & \textbf{Mixlacune} &       & 4     & 2 \\
          & \textbf{Neurophet} &       & 6     & 6 \\
          & \textbf{TeamTea} &       & 2.5   & 4 \\
    \bottomrule
    \end{tabular}%
  \label{tab:dataset_ranking}%
\end{table}%



\subsubsection{Regional variability}
\begin{figure}
    \centering
    \includegraphics[width=0.95\textwidth]{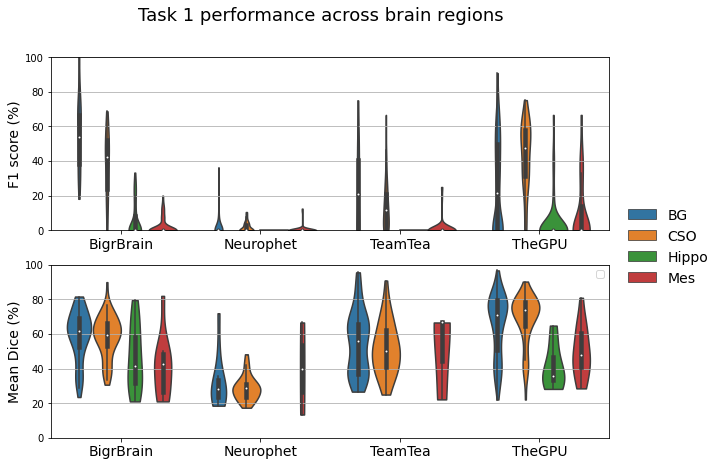}
    \caption{F1 and Mean Dice distribution across the different brain regions for Task 1 - EPVS}
    \label{fig:regional_pvs}
\end{figure}

Metrics variability for Task 1 - EPVS across different brain regions is illustrated for F1 and Mean Dice in Figure \ref{fig:regional_pvs}. 

\subsubsection{Inter-rater variability}
Inter-rater variability was investigated for tasks and datasets for which two raters provided annotation for the same case (Task 1 - EPVS SABRE dataset, Task 3 - Lacunes all datasets) and results are presented in Table \ref{tab:interrater}.

\begin{table}[htbp]
  \centering
  \tiny
  \caption{Metrics values (median [1st quartile - 3rd quartile] presented for the cases where a double rating was available in the test set.}
      \begin{tabular}{ccccccc}
\cmidrule{2-7}          & \multicolumn{3}{c}{\textbf{Detection}} & \multicolumn{3}{c}{\textbf{Segmentation}} \\
          & \multicolumn{1}{l}{\textbf{F1 score R1}} & \multicolumn{1}{l}{\textbf{F1 score R2}} & \multicolumn{1}{l}{\textbf{AED}} & \multicolumn{1}{l}{\textbf{Mean Dice R1}} & \multicolumn{1}{l}{\textbf{Mean Dice R2}} & \multicolumn{1}{l}{\textbf{AVD}} \\
    \midrule
    \multirow{2}[1]{*}{Task 1 - EPVS} & 19.57 & 19.86 & 135.00   & 52.63 & 54.07 & 651.00 \\
          & [13.48 ; 23.81] & [13.58 ; 23.81] & [96.25 ; 316.50] & [52.07 ; 54.51] & [51.43 ; 55.05] & [371.75 ; 2819.75] \\
    \multirow{2}[1]{*}{Task 3 - Lacunes} & 48.45 & 55.84 & 0.50   & 59.03 & 59.49  & 21.51 \\
          & [39.01 ; 61.88] & [0.00 ; 86.36] & [0.00 ; 1.00] & [43.72 ; 64.47] & [44.95 ; 65.88] & [0.00 ; 43.63] \\
    \bottomrule
    \end{tabular}%
  \label{tab:interrater}%
\end{table}%

For Task 1 - EPVS, intra-rater detection was slightly lower than the best method but the inter-rater segmentation performance appeared to be better by quite a strong margin reaching 59.49\% in comparison to the best method at 45.5\%. The detection performance was notably higher for Task 3 - Lacunes with segmentation performance on par with the best performing method.



\subsubsection{Ensembles}
For the creation of EnsembleTop, Task 1 - EPVS used predictions from team TheGPU and BigrBrain, Task 2 - Microbleeds used predictions from MixMicrobleedNet, TeamTea, Zihao, and ValdoNN, while for Task 3 - Lacunes, predictions from Dawai, TeamTea and EMC\_N were used. Table \ref{tab:ensemble} presents the values of the metrics and the corresponding ranking obtained for each type of ensemble (EnsembleAll, the average of all solutions, and EnsembleTop, the average of the top 50\%) across the three tasks. When considering the clinical metrics, performance was higher for both ensemble solutions in Task 1 - EPVS reaching a correlation coefficient of 70.0\% and 74.8\% for EnsembleAll and EnsembleTop respectively for the count and 69.5 and 80.0\% for the volume. For Task 2 - Microbleeds, balanced accuracy was of 77.0\% for EnsembleAll and 79.6\% for EnsembleTop ranking fourth and third compared to all the teams. Finally, for Task 3 - Lacunes, balanced accuracy reached 75.0\% for EnsembleAll, down to 65.3\$ for EnsembleTop slightly lower than the 78.0\% obtained by TeamTea.

\begin{table}[htbp]
 \centering
  \tiny
  \caption{Metrics value presented as median [IQR] for the 4 common metrics across the different ensemble types for the three tasks along with associated ranking}
    \begin{tabular}{cccccc}
    \toprule
          &       & \textbf{F1} & \textbf{AD} & \textbf{Mean Dice} & \textbf{AVD} \\
    \midrule
    \multirow{4}[1]{*}{Task 1 - EPVS} & \multirow{2}[1]{*}{EnsembleAll} & 38.62 [28.10 ; 44.82] & 24.00 [12.00 ; 46.00] & 64.33 [59.14 ; 68.40] & 96.15 [63.67 ; 151.69] \\
          &       & 3.5   & 3.5   & 4     & 2 \\
          & \multirow{2}[0]{*}{EnsembleTop} & 38.86 [31.19 ; 45.13] & 29.00 [15.25 ; 50.25] & 67.38 [58.24 ; 72.23] & 36.10 [20.15 ; 66.33] \\
          &       & 1.5   & 3.5   & 2     & 2 \\
    \multirow{4}[0]{*}{Task 2 - Microbleeds} & \multirow{2}[0]{*}{EnsembleAll} & 66.67 [0.00 ; 100.00] & 1.00 [0.00 ; 1.00] & 81.22 [71.35 ; 87.27] & 12.87 [4.93 ; 27.26] \\
          &       & 4     & 3     & 6.5   & 7 \\
          & \multirow{2}[0]{*}{EnsembleTop} & 75.68 [38.18 ; 100] & 1.00 [0.00 ; 1.00] & 77.90 [29.91 ; 87.23] & 11.25 [2.81 ; 21.82] \\
          &       & 1     & 1     & 3     & 3 \\
    \multirow{4}[1]{*}{Task 3 - Lacunes} & \multirow{2}[0]{*}{EnsembleAll} & 28.57 [0.00 ; 60.87] & 1.00 [0.00 ; 2.00] & 37.98 [22.13 ; 44.55] & 13.05 [0.07 ; 61.03] \\
          &       & 2.5   & 2     & 3.5   & 2 \\
          & \multirow{2}[1]{*}{EnsembleTop} & 30.77 [0.00 ; 66.67] & 1.00 [0.00 ; 2.00] & 38.17 [25.48 ; 45.26] & 9.68 [1.05 ; 63.28] \\
          &       & 2.5   & 2     & 3.5   & 2 \\
    \bottomrule
    \end{tabular}%
  \label{tab:ensemble}%
\end{table}%

\section{Discussion}

This manuscript reports the design and outcome of the \textit{"Where is VALDO?"} challenge that took place as a satellite event of MICCAI 2021. Detection and segmentation of three types of markers of cerebral small vessel disease were evaluated as three distinct tasks namely enlarged perivascular spaces (Task 1), cerebral microbleeds (Task 2) and lacunes (Task 3). Among the 12 distinct participating teams, 9 teams provided a solution for Task 2 and 2 teams competed across all three tasks. 

Although the challenge was designed to address both detection and segmentation aspects, most of the proposed solutions were designed with a segmentation purpose only - the detection performance considered as a by-product of the prediction. This choice may have been influenced partially by the guidelines to provide only the probabilistic segmentation map that was then post-processed to identify the individual connected components instead of requesting instance segmentation and predicted detections as outputs. However, this strategy appeared to generally work well with segmentation performance being on par with detection performance across all three tasks. Interestingly, there was no strong relationship between memory, time expenditure and overall performance with some of the most greedy methods having lower performance than some of the most cost-effective solutions.

Across all tasks, one team proposed a solution not relying on deep-learning and their strategy had the best performance for Task 1 - EPVS possibly because of the fact that EPVS may be relatively easy to characterise in terms of signal and shape signature. However, none of the proposed methods for Task 1 - EPVS made use of the weak annotation data (count on slices). Also, while some methods only used annotated slices, performance may have been lowered by the absence of use of the masks when only specific parts of a given axial slice were annotated (RSS Data). Most deep learning solutions described using a UNet style architecture at one point of their pipeline either as main network for one-stage methods or for the segmentation component for multi-stage solutions. Interestingly, despite four teams describing the use of the nnUNet \citep{isensee2021nnu} architecture for Task 2 - Microbleeds, performance varied greatly across these teams with rank 1, 2, 5 and 7 out of 9. This could potentially be explained by the choice of input data, the dimensionality, or the framework chosen. In the context of microbleeds, using 3D information may be particularly relevant to avoid mimics. This observation highlights the importance of all these steps in the design of a relevant solution, the use of the whole extent of the training data being a key component of the winner's method. Such consideration is particularly relevant when dealing with a modest number of training examples. When considering choices of augmentation, those involving local changes to input images and/or reference annotation (interpolation, intensity changes, spatial deformation) may cause inconsistencies in the case of very small objects of interest. 

In terms of dataset origins, performance was generally higher for the dataset with the highest resolution which was also for Task 2 - Microbleeds and Task 3 - Lacunes the dataset with the highest number of training cases. This is naturally expected as a direct impact on resolution on evaluation metrics and as an overfitting related property. 

The amount of training data (in terms of examples of lesions) appeared also to be relevant when comparing the performance of the methods of Task 1 across the different regions of interest, the regions with the most EPVS (centrum semi-ovale and basal ganglia) being the ones with the highest performance across all methods. This may not only be due to the sheer amount of training data in the remaining regions (hippocampus and mesencephalon) but also to the characteristics of the imaging sequences in these regions and the likelihood for mimics (cysts) and higher variability in presentation. Knowledge of the differences in performance across regions is particularly interesting clinically when associations with risk factors and or clinical function have been made specifically in specific anatomical regions in relation to Alzheimer's Disease \citep{jimenez2018prevalence} and Parkinson's disease \citep{duker2007parkinsonism}.
For Task 1 - EPVS, even for the best teams, the performance presented a large variability which would make their adoption in clinical practice difficult. The overall good correlation between expected and predicted burden may however already be enough to make these tools valuable when investigating associations at population level.
For Task 2 - Microbleeds, it appeared that, when correctly detected, the segmentation of lesions was very good. However, even in the best of teams there were issues at the detection level with both cases missed and cases wrongly considered as containing at least one microbleed. The best teams indicated very few lesions which would be relatively practical to visually inspect and reject if necessary. It is here the absence of a systematic bias towards overcall or undercall could make it difficult to integrate in clinical pipelines.
For Task 3 - Lacunes, performance appeared quite poor on both detection and segmentation metrics, with a general large overcall of lacunes and when detecting them correctly a lower segmentation performance than for Task 2 - Microbleeds. Such performance would require too much time for editing and checking to be adopted in both clinical practice and research studies.

When comparing the performance across all three tasks, it appeared that the performance was higher on tasks for which the variability in element appearance was lower (EPVS with linear shapes and microbleeds with spherical shapes compared to lacunes with more heterogeneous shapes). The metrics investigated as closest to the current clinical measures of interest were generally in agreement with the overall ranking of the challenge but showed stark differences in terms of clinical viability of the proposed solutions. While for Task 1 - EPVS and Task 2 - Microbleeds the proposed solutions achieved reasonable performance in terms of "clinical" metric, only one team performed reasonably well for Task 3 - Lacunes, with all other solutions systematically finding many lacunes even when there were none. This may be due to the large variability in appearance (i.e. shape, location, intensity signature) as well as the lower number of examples of this type of lesions when compared to those of Task 1 - EPVS and Task 2 - Microbleeds. 
With all solutions generally producing many false positives, the time required to go through each case and reject many wrongly detected lesion candidates would be prohibitive for clinical adoption. One must however keep in mind that none of these solutions were optimized for this metric and may have performed differently otherwise. In this case the addition of auxiliary tasks in the learning framework to abide to a priori knowledge of burden distribution or to directly optimize such metrics may have interesting results. 

In a field where adequate research biomarkers have yet to be properly defined and proven to be reliable \citep{smith2019harmonizing}, these observations regarding clinical metrics may lead to define different tasks and solutions for the targeted markers according to their purpose: clinical practice or research. While location, individual volume and shape information may become of interest in the research context as potential new biomarkers, thereby highlighting segmentation as an interesting end-goal, these characteristics may not be yet relevant in the clinical context. In clinical practice, one could imagine a two-stage pipeline with 1) whole-image level classification favouring sensitivity for the flagging of scans where an assessment is required for the presence or absence of a specific marker 2) Specification of lesion location (if needed) for the scans that have been flagged as containing a marker. This second step may be particularly relevant when supporting diagnosis (e.g., distinction between amyloid angiopathy and hypertensive pathology according to microbleed location) or to the explanation of the clinical presentation (e.g., lacune on crucial white matter tract). 

 A key aspect, not measured here, is the ability of the proposed methods to be used in clinical settings with scans likely to be of lower resolution and to have more artefacts as well as present simultaneously other markers of pathology (e.g stroke, tumours). With the continuous progress in acquisition protocols and the democratization of scanning abilities, research-grade scanning protocols such as those used in this challenge may become available routinely, thereby limiting issues of protocol related generalizability.
However, cohort-related bias may be more difficult to overcome. In fact, in the challenge, data came only from population cohorts and did not include patients with dementia as would be frequent in memory clinics. While efforts were made to provide training examples from the whole spectrum of lesion burden, specific pathological presentations may be missing and the generalizability of the proposed solutions would need to be assessed in these contexts.

\paragraph{Conclusion}
In this challenge assessing the current segmentation and detection performance of three markers of cerebral small vessel disease, namely EPVS, Microbleeds and Lacunes, methods targeting directly the segmentation were often quite successful in detecting these small structures. Number of elements on which to train the solutions was strongly predictive of performance, both across tasks and regionally. Manually engineered features became in the case of EPVS relevant enough to compete with deep-learning based strategies. Strikingly, all the presented methods proposed a training based on dense labelling, discarding the weak labelling available for Task 1 - EPVS. While for Task 1 - EPVS and Task 2 - Microbleeds some demonstrated they could potentially be used for population-based research, the large variability in performance across cases may require lengthy visual censoring if they were to be used for individual cases. In this context, it could be relevant to further include the evaluation of performance variability in the assessed tasks. In addition, systematic assessment of prediction confidence (as proposed with the uncertainty metrics of Task 3 - Lacunes) would be of interest for the design of practical implementation.      

\paragraph{Funding}
The challenge prizes were provided by Nvidia and Icometrix.
The SABRE study was funded at baseline by the Medical Research Council, Diabetes UK, and British Heart Foundation and at follow-up by the Wellcome Trust (082464/Z/07/Z), British Heart Foundation (SP/07/001/23603, PG/08/103, PG/12/29/29497 and CS/13/1/30327) and Diabetes UK(13/0004774).  
The Rotterdam Scan Study is supported by the Erasmus MC University Medical Center, the Erasmus University Rotterdam, the Netherlands Organization for Scientific Research (NWO) Grant 918-46-615, the Netherlands Organization for Health Research and Development (ZonMW), the Research Institute for Disease in the Elderly (RIDE), and the European Union Seventh Framework Programme (FP7/2007–2013) under grant agreement No. 601055, VPH-DARE@IT, the Dutch Technology Foundation STW (Perspectief programme: Population Imaging Genetics
The ALFA study is supported by the La Caixa Foundation.
CHS is funded by an Alzheimer's Society Junior Fellowship (AS-JF-17-011). KVW and SC are supported by the Deep Learning for Medical Image Analysis (DLMedIA) (project no. P15-26), funded by the Dutch Technology Foundation STW, which is part of the Netherlands Organisation for Scientific Research (NWO) and which is partly funded by the Ministry of Economic Affairs, with co-financing by Quantib. FD was funded by Netherlands Organisation for Health Research and Development 104003005. 
BM, BW and FK are supported through the SFB 824, subproject B12, supported by Deutsche Forschungsgemeinschaft (DFG) through TUM International Graduate School of Science and Engineering (IGSSE), GSC 81. IE is funded by DComEX (Grant agreement ID: 956201). BM acknowledges support by the Helmut Horten Foundation. MdG is an employee of, and holds shares in GSK. GSK had no role in the design of this challenge, or the interpretation of the results. MdB is supported by Netherlands Organisation for Scientific Research (NWO) project VI.C.182.042.
SI and LL have received funding from the Innovative Medicines Initiative 2 Joint Undertaking under Amyloid Imaging to Prevent Alzheimer’s Disease (AMYPAD) grant agreement No. 115952 and European Prevention of Alzheimer’s Dementia (EPAD) grant No. 115736. This Joint Undertaking receives the support from the European Union’s Horizon 2020 Research and Innovation Programme and EFPIA.
HJK was supported by the Galen and Hilary Weston Foundation under the Novel Biomarkers 2019 scheme (\textnumero UB190097). JLM is currently a full-time employee of Lundbeck and has served previously as a consultant or on advisory boards for the following for-profit companies, or has given lectures in symposia sponsored by the following for profit companies: Roche Diagnostics, Genentech, Novartis, Lundbeck, Oryzon, Biogen, Lilly, Janssen, Green Valley, MSD, Eisai, Alector, BioCross, GE Healthcare, and ProMIS Neurosciences. JDG is supported by the Spanish Ministry of Science and Innovation (RYC-2013-13054), has received research support from GE Healthcare, Roche Diagnostics and Hoffmann-La Roche and speaker's fees from Biogen and Philips.

\paragraph{Acknowledgements}
We are particularly thankful to all participants of the ALFA, RSS and SABRE study. We also would like to thank the team of GrandChallenge.org for their technical support and guidance.
The ALFA group study is composed of Müge Akinci, Eider M Arenaza-Urquijo, Annabella Beteta, Anna Brugulat-Serrat, Raffaele Cacciaglia, Alba Cañas, Irene Cumplido, Carme Deulofeu, Ruth Dominguez, Maria Emilio, Carles Falcon, Karine Fauria, Sherezade Fuentes, José Maria González de Echavarri-Gómez, Oriol Grau-Rivera, Laura Hernandez, Gema Huesa, Jordi Huguet, Paula Marne, Marta Milà-Alomà, Tania Menchón, Carolina Minguillon, Arcadi Navarro, Grégory Operto, Eva M Palacios, Eleni Palpatzis, Cleofé Peña-Gómez, Albina Polo, Sandra Pradas, Blanca Rodríguez-Fernández, Aleix Sala-Vila, Gonzalo Sánchez-Benavides, Gemma Salvadó, Mahnaz Shekari, Anna Soteras, Laura Stankeviciute, Marc Suárez-Calvet, Marc Vilanova, Natalia Vilor-Tejedor.

\bibliography{sn-bibliography}


\end{document}